\documentclass[11pt]{article}

\usepackage[final]{acl}

\usepackage{times}
\usepackage{latexsym}

\usepackage[T1]{fontenc}

\usepackage[utf8]{inputenc}

\usepackage{microtype}

\usepackage{inconsolata}

\usepackage{graphicx}

\usepackage{hyperref}       %
\usepackage{url}            %
\usepackage{booktabs}       %
\usepackage{amsfonts}       %
\usepackage{nicefrac}       %
\usepackage{xspace}
\usepackage{lipsum}
\usepackage{amsmath}
\usepackage{enumitem}
\usepackage{trimclip}

\usepackage[table]{xcolor}
\usepackage[dvipsnames]{xcolor}

\usepackage{listings}
\usepackage{xcolor}

\definecolor{jsonstring}{HTML}{0057AE}   %
\definecolor{jsonnumber}{HTML}{B08000}   %
\definecolor{jsonbrace}{HTML}{644A9B}    %
\definecolor{jsonbracket}{HTML}{006E28}  %
\definecolor{jsonbool}{HTML}{0057AE}     %
\definecolor{jsonpunct}{HTML}{4D4D4D}    %
\definecolor{bgcolor}{HTML}{FAFAFA}      %

\lstdefinelanguage{json}{
    basicstyle=\ttfamily\tiny,
    numbers=left,
    numberstyle=\tiny\color{gray},
    stepnumber=1,
    numbersep=8pt,
    showstringspaces=false,
    breaklines=true,
    breakatwhitespace=true,
    frame=single,
    rulecolor=\color{lightgray},
    backgroundcolor=\color{bgcolor},
    keywords={true,false,null},
    keywordstyle=\color{jsonbool}\bfseries,
    literate=
       *{0}{{{\color{jsonnumber}0}}}{1}
        {1}{{{\color{jsonnumber}1}}}{1}
        {2}{{{\color{jsonnumber}2}}}{1}
        {3}{{{\color{jsonnumber}3}}}{1}
        {4}{{{\color{jsonnumber}4}}}{1}
        {5}{{{\color{jsonnumber}5}}}{1}
        {6}{{{\color{jsonnumber}6}}}{1}
        {7}{{{\color{jsonnumber}7}}}{1}
        {8}{{{\color{jsonnumber}8}}}{1}
        {9}{{{\color{jsonnumber}9}}}{1}
        {:}{{{\color{jsonpunct}{:}}}}{1}
        {,}{{{\color{jsonpunct}{,}}}}{1}
        {\{}{{{\color{jsonbrace}\{}}}{1}
        {\}}{{{\color{jsonbrace}\}}}}{1}
        {[}{{{\color{jsonbracket}[}}}{1}
        {]}{{{\color{jsonbracket}]}}}{1},
    string=[b]{"},
    stringstyle=\color{jsonstring},
    comment=[l]{\#},
    commentstyle=\color{jsonstring}\itshape,    
}
\usepackage{caption}
\definecolor{d3red}{HTML}{d62728}
\definecolor{d3green}{HTML}{2ca02c}

\definecolor{posA}{HTML}{F3FAF3}
\definecolor{posB}{HTML}{EAF6EA}
\definecolor{warnA}{HTML}{FFFBEA}
\definecolor{negA}{HTML}{FFF1F1}

\newcommand{\syscite}[2]{\textsc{#1}~\cite{#2}}

\title{%
The Scientific Contribution Graph:\\Automated Literature-based Technological Roadmapping at Scale}

\author{
 \textbf{Peter A. Jansen\textsuperscript{1,2}}
\\ 
 \textsuperscript{1}University of Arizona, 
 \textsuperscript{2}Allen Institute for Artificial Intelligence\\  
 \texttt{pajansen@arizona.edu}
}

\begin{document}
\maketitle

\begin{abstract}
Scientific contributions rarely develop in isolation, but instead build upon prior discoveries. We formulate the task of automated technological roadmapping as extracting scientific contributions from scholarly articles and linking them to their prerequisites. We present the \textsc{Scientific Contribution Graph}, a large-scale resource containing 6 million detailed scientific contributions extracted from 655k open-access papers spanning computer science, medicine, biology, physics, chemistry, and other sciences, and connected by 36 million prerequisite edges. We further introduce scientific prerequisite prediction, a scientific discovery task in which models predict which existing technologies can enable future discoveries, and show that contemporary models are rapidly improving on this task, reaching 0.48 MAP when evaluated using temporally-filtered backtesting. We anticipate technological roadmapping resources such as this will support scientific impact assessment and automated scientific discovery.\footnote{Code, API, data, demo: \url{https://github.com/cognitiveailab/scientific-contribution-graph}}
\end{abstract}

\section{Introduction}

Sir Isaac Newton famously wrote, ``If I have seen further, it is by standing on the shoulders of giants''. Scientific contributions are rarely developed in isolation, but build upon prior contributions, such as problem framings, experimental methods, and empirical findings. Understanding these prerequisite relationships is important for studying scientific progress, and for automated scientific discovery systems that must reason about which existing capabilities can be used to develop new ones \cite[e.g.][]{lu2024aiscientistfullyautomated,jansen-etal-2025-codescientist,baek-etal-2025-researchagent}.

Currently, the way scientific contributions build upon one another is typically represented using either paper-level citations or sentence-level triples extracted from papers. Citation analysis captures paper-level links \cite{Wang2022DisenCiteGDA,Zhang2021MeasuringAEA,ammar-etal-2018-construction}, but citations often provide background context \cite[e.g.][]{li-ouyang-2024-related,li-etal-2022-corwa}, rather than indicating that one work was directly built upon another. Impactful-citation detection methods \cite[e.g.][]{Valenzuela2015IdentifyingMC,Nazir2020ImportantCIA} partially control for this, but still operate at a coarse, whole-paper level: they cannot resolve, for example, that the \textit{data} developed in Paper A depends specifically upon the \textit{problem formulation} proposed in Paper B. Conversely, scientific information extraction (IE) systems often extract sentence-level entities or triples, such as \textsc{$<$speech enhancement, uses, spectral restoration method$>$} \cite[a triple from \textsc{CS-KG};][]{dessi2025cs}. These triples are often decontextualized: it is unclear what specific technique is being discussed, or whether, and how, one contribution builds upon another. This leaves a representational utility gap between paper-level citation graphs and sentence-level triples.

\begin{table*}[t!]
\centering
\footnotesize
\setlength{\tabcolsep}{3pt}
\begin{tabular}{p{0.29\linewidth}p{0.12\linewidth}p{0.21\linewidth}p{0.20\linewidth}p{0.11\linewidth}}
\toprule
\textbf{System} & \textbf{Input} & \textbf{Extraction Task} & \textbf{Dependency Model} & \textbf{Scale} \\
\midrule
\rowcolor[HTML]{F3F3F3}
\syscite{SciERC}{luan-etal-2018-multi}
& 500 abstracts
& Span/relation labeling
& Typed triples
& 4.7k triples \\

\syscite{SciREX}{jain-etal-2020-scirex}
& 438 papers
& Span/relation labeling
& N-ary tuples
& 2.2k tuples \\
\rowcolor[HTML]{F3F3F3}
\syscite{NLPCont.}{dsouza-etal-2021-semeval}
& 442 papers
& Span/relation labeling
& General SPO triples
& 30k triples \\

\syscite{SciNLP-KG}{mondal-etal-2021-end}
& 30k papers
& Span/relation labeling
& Typed triples
& 8.1k triples \\
\rowcolor[HTML]{F3F3F3}
\defcitealias{magnusson-friedman-2021-extracting}{Magnusson et al.}
\textsc{SciClaim} (\citetalias{magnusson-friedman-2021-extracting},~\citeyear{magnusson-friedman-2021-extracting})
& 901 sentences
& Span/relation labeling
& Sentence-level graphs
& 5.3k triples \\

\syscite{CS-KG V2}{dessi2025cs}
& 14.5M papers
& Span/relation labeling
& Typed triples 
& 67.5M triples \\

\midrule
\rowcolor[HTML]{E3E3E3}
\begin{tabular}[t]{@{}l@{}}\textbf{\textsc{Scientific Contribution}}\\[-1pt] \textbf{\textsc{Graph} (This work)}\end{tabular}
& 655k papers
& \begin{tabular}[t]{@{}l@{}}\textbf{Seq2Seq contribution}\\[-1pt]\textbf{generation on full text}\end{tabular}
& \begin{tabular}[t]{@{}l@{}}\textbf{Prereq. extraction +}\\[-1pt]\textbf{cross-paper alignment}\end{tabular}
& \textbf{6M nodes / 36M edges} \\
\bottomrule
\end{tabular}
\vspace{-2mm}
\caption{\footnotesize A comparison of existing scientific knowledge graph efforts, including input source, extraction task framing, dependency model for edge construction, and scale. Prior systems primarily extract typed triples over short entity spans, where this work generates detailed scientific contribution nodes through seq2seq modeling, and edges through cross-paper contribution alignment.}
\label{tab:related-work-comparison} %
\vspace{-2mm}
\end{table*}

This work addresses this gap with the \textsc{Scientific Contribution Graph}, a large-scale resource that represents scientific progress at an intermediate granularity: finer than papers, but richer than triples. Nodes represent individual scientific contributions, while edges represent how one contribution builds upon precursor contributions. The graph contains 6M contributions from 655k open-access papers seeded from the \textsc{ACL Anthology}, and linked by 36M prerequisite edges, with an average of 9 extracted contributions per paper.

Rather than representing nodes and edges as short entity mentions or triples, contribution nodes contain generated names and detailed summary descriptions, while edges explain how the source text describes one contribution as building upon another. Where nodes in existing graphs such as \textsc{CS-KG} average approximately 3 tokens, nodes and edges in the \textsc{Scientific Contribution Graph} contain an average of 130 and 71 tokens, respectively. Building this detailed graph requires modeling knowledge graph construction as open-ended sequence-to-sequence extraction over paper full text, rather than span labeling or triple extraction over sentences or paragraphs \cite{dagdelen2024structured,shamsabadi-etal-2024-large}. This is computationally nontrivial: the pipeline requires an average of 43 LLM calls per paper to extract contributions, identify prerequisites, and align them to graph contributions. To demonstrate utility for automated scientific discovery, this work introduces a scientific prerequisite prediction task, where a model must identify existing technologies required to develop a novel target technology. Using temporally filtered  backtesting \cite[e.g.][]{luo-etal-2018-scientific,jansen-etal-2025-matter}, we empirically demonstrate contemporary language models have recently progressed from near-chance to moderate performance, suggesting near-term utility for scientific discovery.

The contributions of this work are:
\begin{enumerate}[itemsep=0pt, topsep=2pt, leftmargin=*]
    \item The \textsc{Scientific Contribution Graph}, a large-scale resource for technological roadmapping containing 6M detailed scientific contributions from 655k open-access papers, linked by 36M prerequisite edges describing how technologies build upon precursor technologies.
    \item The formulation of scientific prerequisite prediction as a technological requirement prediction task for automated scientific discovery, paired with an empirical demonstration showing that contemporary language models have rapidly improved on this task, progressing from chance performance to 0.48 \textsc{MAP} in the last two years.
\end{enumerate}

\begin{figure*}[t!]
  \centering  
  \includegraphics[scale=0.93]{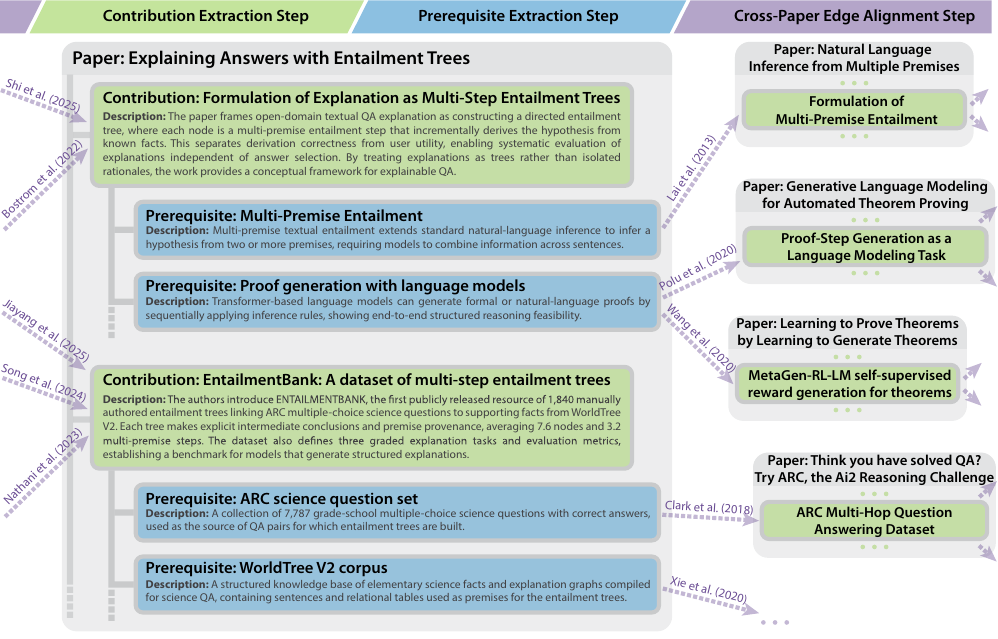}
  \vspace{-4mm}
  \caption{\footnotesize Illustration of the graph construction pipeline, including: (1) extracting contributions from papers, (2) extracting prerequisites for each contribution, and (3) creating contribution dependency edges by aligning prerequisites to contributions from cited papers. Due to space limitations, only a small subset of contributions, prerequisites, edges, and content are shown (see Table~\ref{tab:impact_entailmentbank} for a list of contributions from this example paper (names only), and \textsc{Appendix}~\ref{sec:example-contributions} for a full example). On average 9 contributions are extracted from each paper, each containing 6 prerequisite edges.}
  \label{fig:system_overview}
  \vspace{-3mm}
\end{figure*}

\section{Related Work}

{\flushleft\textbf{Extracting Scientific Contributions:}} Prior work has extracted scientific contributions from literature, most often as span-labeled named-entity recognition (NER) or structured information extraction.  NER systems typically pre-define 3 to 10 entity classes, such as \textsc{datasets}, \textsc{models}, \textsc{methods}, \textsc{tasks}, and \textsc{metrics} \cite{Das2025WhatsIY,pramanick-etal-2025-nature,otto-etal-2023-gsap,dsouza-etal-2021-semeval,luan-etal-2018-multi}, then identify contributions in abstracts or full text using classification or span-labeling. Recent work models scientific IE as sequence-to-sequence (seq2seq) extraction: Shamsabadi et al.~\shortcite{shamsabadi-etal-2024-large} extract structured schemas for virology papers, including the virus studied, location, time, and reproduction rate, while Dagdelen et al.~\shortcite{dagdelen2024structured} extract material data using structured schemas. This seq2seq IE framing is used at scale here to build the knowledge graph.

{\flushleft\textbf{Scientific Knowledge Graphs:}} Existing scaled scientific knowledge graphs are primarily represented as triples or N-ary relations, as shown in Table~\ref{tab:related-work-comparison}.  The largest resource, CS-KG V2 \cite{dessi2025cs}, is extracted using the SCICERO \cite{Dess2022SCICEROADA} model, which maps entities to a standard ontology. Mondal et al.~\shortcite{mondal-etal-2021-end} extract two evaluation-centered relations, \textsc{EvaluatedOn/By}, from 30k NLP-domain papers; SciClaim \cite{magnusson-friedman-2021-extracting} extracts sentence-level claim graphs; and D'Souza et al.~\shortcite{dsouza-etal-2021-semeval} extract open-ended triples from 442 papers, labeling them with 12 broad contribution types. In contrast, this work models extraction as open-ended sequence-to-sequence generation, extracting contribution names and detailed descriptions from full-text, rather than entity labels or triples from sentences using a predefined schema. These contributions are then linked to prerequisite contributions mentioned in the source text using an alignment procedure, yielding a large and detailed graph of 6M contributions and 36M edges extracted from 655k papers.

{\flushleft\textbf{Automated Scientific Discovery:}} Literature-based discovery is often framed as identifying ``undiscovered public knowledge'', or plausible relations not explicitly stated in the literature, but inferred by combining published relations \cite{Swanson1986FishOil,swanson1988migraine}. Modern work often formulates discovery as link prediction over graphs of scientific concepts \cite{cesario2024survey,pu2023graph,crichton2020neural}, or uses literature to generate research directions through iterative refinement \cite{si2025can}, literature-generated idea facets \cite{radensky2024scideator}, or extrapolation from literature chains \cite{vasu-etal-2025-hyper,li-etal-2025-chain-ideas}. Most related, \textsc{Giants} \cite{he2026giants} tests whether LLMs can ideate future insights from two component technologies, while \textsc{CodeScientist} \cite{jansen-etal-2025-codescientist} ideates and tests literature-derived technologies through code-based experimentation. This work instead focuses on technological roadmapping -- representing how scientific contributions build upon prerequisite contributions -- and evaluating the reverse prediction problem of predicting which existing scientific contributions are required to develop a novel target technology.

\section{Problem Formulation}

We formulate technological roadmapping as an automatic knowledge graph construction problem.  Let $D$ denote the input document corpus, instantiated here as scientific papers.  The synthesis process maps $D$ to a contribution graph ($G$): $D \rightarrow G$.  The graph $G$ consists of a set of nodes $c_i \in C$, each representing an individual scientific contribution, and a set of edges $e_j \in E$ that link scientific contributions to their prerequisite technologies.

Each contribution $c_i$ is represented as a \textsc{<name, description, metadata>} tuple, $\langle n_i, d_i, m_i \rangle$, where $n_i$ is the name of the scientific contribution, $d_i$ is a natural-language description of the contribution, and $m_i$ is an optional set of associated metadata, such as the title of the source paper and sections of the source paper that mention the contribution.  Each edge $e_j$ is represented as a \textsc{<precursor node, dependent node, attributes>} tuple, $\langle c_{\mathrm{pre}}, c_{\mathrm{dep}}, a_j \rangle$, where $c_{\mathrm{pre}}$ and $c_{\mathrm{dep}}$ are the precursor and dependent contribution nodes, respectively, and $a_j$ is an optional set of edge attributes generated during extraction, such as a natural-language explanation of why $c_{\mathrm{pre}}$ is described as a prerequisite technology for $c_{\mathrm{dep}}$.

\subsection{Edge Alignment}

Papers typically describe how new scientific contributions build on existing contributions using a combination of references and prose.  For example, the toy sentence \textit{``Our Contrastive-RLHF model modifies the preference-weighted RL training loss of Smith et al. (2025) with a contrastive objective...''} describes the dependent contribution (\textit{``Contrastive-RLHF model''}), a candidate prerequisite technology (\textit{``preference-weighted RL training loss''}), and its source (\textit{``Smith et al. 2025''}).  We model edge construction as a two-step process.  First, for each dependent contribution $c_{\mathrm{dep}}$, we extract a list of candidate prerequisite technologies $P(c_{\mathrm{dep}})$.  Each candidate prerequisite $p_k \in P(c_{\mathrm{dep}})$ is represented as a \textsc{<name, description, explanation, reference>} tuple, $\langle pn_k, pd_k, pe_k, pr_k \rangle$, where $pn_k$, $pd_k$, and $pe_k$ describe the prerequisite technology name, description, and a short natural-language explanation of why it is a prerequisite.  The reference $pr_k$ typically identifies a source paper, including its title, year, and first author, but may also be an internal reference, when a later contribution in a paper depends upon an earlier contribution in the same paper, or an external artifact reference, such as a tool or software package referenced by a \textsc{URL}.

Second, we align each candidate prerequisite $p_k$ to a concrete precursor contribution node $c_{\mathrm{pre}}$.  To generate an edge $e_j$ linking a dependent contribution $c_{\mathrm{dep}}$ to a precursor contribution, an aligner $A$ takes as input the dependent contribution $c_{\mathrm{dep}}$, the candidate prerequisite $p_k$, and the paper introducing the prerequisite, $d_{\mathrm{pre}}$, and produces an edge $e_j$ as output: $e_j = A(c_{\mathrm{dep}}, p_k, d_{\mathrm{pre}})$.  The alignment algorithm used in this work is described below.

\section{Knowledge Graph Construction}

Figure~\ref{fig:system_overview} summarizes the construction pipeline. %
Starting from an initial corpus of papers, the pipeline iteratively expands the contribution graph by extracting contribution nodes, identifying prerequisite technologies, and aligning those prerequisites to source contributions.  After each batch, the next papers are selected from cited papers not yet in the graph, prioritized by citation frequency within the graph to minimize edge sparsity.

\subsection{Corpus of Open Access Papers}
The \textsc{Scientific Contribution Graph} currently contains contributions extracted from \textsc{655k} open-access papers drawn from two permissively licensed sources.  The seed corpus is a crawl of the \textsc{ACL Anthology}\footnote{\url{https://aclanthology.org/}}, containing approximately \textsc{120k} papers licensed under Creative Commons \textsc{CC-BY} or \textsc{CC-BY-SA} licenses.  Additional papers are drawn from the Semantic Scholar Open Research Corpus \textsc{(S2ORC)} \cite{lo-etal-2020-s2orc}, which provides full text for open-access scientific papers. The fields of study of papers are shown in Table~\ref{tab:s2_fields_of_study}.

\subsection{Extraction Pipeline}

The extraction pipeline proceeds in four stages:

{\flushleft\textbf{Step 1: Paper Selection:}} At each iteration, the crawler selects source papers $d_{\mathrm{pre}}$ needed by the aligner to convert currently unresolved prerequisite references into graph edges.  It creates a histogram over unresolved $d_{\mathrm{pre}}$ references: source papers cited by existing candidate prerequisites, but not yet included in the graph.  The most frequently cited open-access papers are selected for the next batch.  For \textsc{ACL Anthology} papers, or when full text is unavailable in \textsc{S2ORC}, paper text is extracted from the PDF using \textsc{Marker}\footnote{\url{https://github.com/datalab-to/marker}}.

{\flushleft\textbf{Step 2: Extracting Contributions (Nodes):}} Contribution extraction is modeled as a sequence-to-sequence task \cite{sutskever2014sequence}, taking paper full text as input and returning \textsc{JSON}-formatted contribution nodes.  For each paper, an instruction-formatted prompt extracts a set of contributions, specifying task desiderata, the desired level of extraction detail, and 17 high-level contribution categories, including \textit{new problem formulations}, \textit{theoretical insights}, \textit{conceptual frameworks}, \textit{resources}, \textit{analyses}, \textit{tools}, \textit{metrics}, \textit{models}, and \textit{algorithms}, with the full list shown in Table~\ref{tab:contribution_categories} in the \textsc{Appendix}.  Each extracted contribution includes its name, description, contribution type(s), and the paper sections where the information is found.  On average, each paper includes 9.1 scientific contributions.

{\flushleft\textbf{Step 3: Extracting Candidate Prerequisites:}} For each contribution extracted in Step 2, an instruction-formatted prompt extracts candidate prerequisite contributions from the paper full text, using both reference context and prose descriptions of how the contribution builds on prior work.  Each candidate prerequisite is represented as a \textsc{<name, description, explanation, reference>} tuple, where the reference identifies the source needed for later edge alignment.  References may be \textit{paper references} to contributions from other papers, \textit{internal references} to earlier contributions in the same paper, or \textit{artifact references}, typically software or tools cited through \textsc{URL}s rather than paper citations.

\begin{table*}[t!]
\centering
\footnotesize
\setlength{\tabcolsep}{4pt}
\begin{tabular}{p{0.24\linewidth}p{0.40\linewidth}p{0.17\linewidth}cc}
\toprule
\textbf{Alignment Category} & \textbf{Interpretation} & \textbf{Symbol} & \textbf{Count} & \textbf{Prop.} \\
\midrule
\rowcolor{posA}
Direct Matches & Reference set and model identify same contribution & $R \rightarrow M$ & 99 & 38\% \\
\rowcolor{posA}
Additional Valid Contributions & Valid model contribution not in reference set & $M \rightarrow \text{unmatched}$ & 64 & 25\% \\
\rowcolor{posA}
High-Resolution Decomp. & Model refines one reference contribution into multiple contributions & $R \rightarrow (M_1,\ldots,M_n)$ & 50 & 19\% \\
\rowcolor{warnA}
Coarser Decomposition & Model combines multiple reference contributions into one contribution & $(R_1,\ldots,R_n) \rightarrow M$ & 24 & 9\% \\
\rowcolor{negA}
Missed Contributions & Reference-set contribution missed by model & $R \rightarrow \text{unmatched}$ & 22 & 9\% \\
\midrule
\rowcolor[HTML]{F3F3F3}
Total &  &  & 258 & 100\% \\
\bottomrule
\end{tabular}
\vspace{-2mm}
\caption{\footnotesize Alignment between reference-set (R) and model-based (M) scientific contribution extraction, broken down into 5 alignment categories. The model achieves 91\% recall of reference-set contributions, while also identifying 25\% additional valid contributions not captured in the manual reference set. In many cases (19\%), the model produces higher-resolution decompositions, breaking coarser reference contributions into finer-grained contributions.}
\label{tab:contribution-alignment}
\vspace{-2mm}
\end{table*}

{\flushleft\textbf{Step 4: Edge Alignment Across Contributions:}} After Step 3, each extracted contribution $c_{\mathrm{dep}}$ is paired with candidate prerequisites $p_k$, but these are not yet graph edges: each $p_k$ must first be aligned to one or more concrete precursor contribution nodes $c_{\mathrm{pre}}$ from its referenced source paper $d_{\mathrm{pre}}$.  For example, if a paper contributes ``formalizing multi-step entailment trees'' and cites ``multi-premise entailment'' from \textit{Natural Language Inference from Multiple Premises} as a prerequisite, the aligner searches the contributions extracted from that source paper and identifies the best-matching precursor contribution, such as ``multi-premise entailment task'' (see Figure~\ref{fig:system_overview}).  The aligner then creates an edge from the precursor contribution to the dependent contribution, indicating that \textit{entailment trees} build on \textit{multi-premise entailment}.  As in earlier steps, alignment is implemented as an instruction-formatted language-model call.  The prompt includes task desiderata, the dependent contribution $c_{\mathrm{dep}}$, the candidate prerequisite $p_k$, and all extracted contributions from $d_{\mathrm{pre}}$, where the model selects the contribution(s) that best align.  Alignment is modeled as one-to-many because a single prerequisite reference may correspond to multiple concrete contributions in the source paper.

{\flushleft\textbf{Extraction Model and Cost:}} All extraction stages use \textsc{GPT-OSS-120B}, chosen for its long-context performance and permissive open weights. To maximize quality, batch processing is not used; contributions and alignments are processed individually where possible. Each paper therefore requires many LLM calls: one for contribution extraction, 9.1 on average for prerequisite extraction, and one per prerequisite edge for alignment, with 6.1 prerequisites per contribution on average (61\% from other papers; 34\% internal, and 5\% artifacts). In practice, this can exceed 40 calls per paper, at an estimated cost of \$0.10 per paper at current rates.

{\flushleft\textbf{Search Embeddings:}} To support fast search, the \textsc{API} includes \textsc{Qwen3-Embedding-0.6B} embeddings for all 6M contribution nodes, generated from concatenated contribution names and descriptions.

\subsection{Evaluation}
\label{sec:evaluation}

A targeted validation was conducted to measure adherence to contribution and prerequisite extraction desiderata.  Reference sets were constructed according to these desiderata and compared to model-generated extractions.  Results, detailed below, show high overall precision and recall for both scientific contribution and prerequisite extraction.

\subsubsection{Contribution Extraction}

Contribution extraction was evaluated using a manual reference set constructed according to the same desiderata in the extraction prompt, including granularity and specificity requirements.  Across 25 papers, a domain expert (the system designer) identified 163 contributions, while the extraction pipeline identified 226 contributions.  As extracted contributions may differ in specificity -- a single coarse contribution may correspond to several more atomic contributions, and vice versa -- evaluation is modeled as a many-to-many alignment problem.  The five alignment classes and their proportions are shown in Table~\ref{tab:contribution-alignment}.

For each paper, model-extracted contributions were aligned to the manually identified contributions and assigned one of five alignment classes: direct match, higher-resolution decomposition, coarser decomposition, additional valid contribution, or missed contribution.  Model contributions without a corresponding manual contribution were reviewed and counted as additional valid contributions when they satisfied the extraction desiderata.  Manually identified contributions with no aligned model contribution were counted as missed contributions.  Proportions are calculated over the union of valid model and manual contributions.

The results in Table~\ref{tab:contribution-alignment} show overall strong performance: 91\% of manually identified contributions were successfully extracted by the model, while 9\% were missed.  The model also identified many additional valid contributions (25\%), most often secondary contributions such as follow-on experiments or analyses missed during manual labeling, and frequently produced higher-resolution decompositions of coarser manual annotations (19\%).  To further validate these quality measurements, a second annotator (a domain expert in natural language processing and uninvolved with this work) was paid to perform the same task on 25 papers of their choosing, with the results of both annotators shown in Table~\ref{tab:contribution-alignment-both} (see \textsc{Appendix}).  The results mirror those of the first domain expert: 94\% of manually identified contributions were successfully extracted, while 6\% were missed, and many additional contributions (33\%) were found by the model and labeled as correct by the second domain expert.  Both domain experts reported the extractor qualitatively appeared very strong at the task, and performing fine-grained decomposition, with the exception of a small number of misses. Taken together, these results indicate high overall performance at extracting scientific contributions.

\subsection{Prerequisite Identification}

Prerequisite identification quality was evaluated manually using the same desiderata specified in the extraction prompt.  For 25 scientific contributions, the domain expert identified 91 prerequisite contributions by examining the source paper text, then compared these manually identified prerequisites to the 149 prerequisites generated by the extraction pipeline.  
To avoid missing prerequisites, the prompt encourages high recall while also distinguishing between \textit{core} or direct prerequisites and valid but less direct \textit{peripheral} prerequisites.
Of the 149 prerequisites identified by the extraction pipeline, 96 were labeled by the pipeline as core prerequisites and 53 as peripheral prerequisites. 

Each model-extracted prerequisite was manually labeled as \textit{core}, \textit{peripheral}, or \textit{incorrect} by the domain expert.  Manual evaluation labeled 93\% of prerequisites labeled \textit{core} by the pipeline as \textit{core}, increasing to 97\% when \textit{peripheral} prerequisites are also counted as valid.  Prerequisites labeled \textit{peripheral} by the pipeline were most often also judged peripheral: 58\% were labeled \textit{peripheral}, and 87\% were labeled either \textit{core} or \textit{peripheral}. Only 7\% of extracted prerequisites (10 of 149) were labeled \textit{incorrect}, primarily cases where the candidate prerequisite was only weakly related or where the dependency direction was reversed.  The manual evaluation identified 9 additional missed prerequisites not recovered by the pipeline, suggesting high recall (94\%) for prerequisite extraction. 

\subsection{Edge Alignment}

\begin{table}
  \footnotesize
  \centering
  \begin{tabular}{lccc}
  \toprule
  ~ & \multicolumn{3}{c}{\textbf{Edge Confidence Rating}}\\
  \textbf{Judge Model} & \textbf{high} & \textbf{low} & \textbf{all} \\
  \midrule
  \textsc{GPT-5.5}         &  0.7\% &  9.4\% &  4.4\% \\
  \textsc{Claude Sonnet 5} &  8.3\% & 17.0\% & 12.0\% \\
  \midrule
  \textsc{\# Samples}      &  144   &  106   &  250   \\
  \bottomrule
  \end{tabular}
  \vspace{-2mm}
  \caption{\footnotesize Assessed error rates on edges by two different \textsc{LLM-as-a-judge} models, estimating overall edge alignment success between 88-96\%.  Edge confidence ratings (high, low) are provided by the extraction model as additional metadata in the graph, and allow filtering to only high-confidence edges. \label{tab:edge-match-confidence}}
  \vspace{-2mm}
\end{table}
  
We evaluated the edge alignment model by measuring how often a candidate prerequisite $p_k$ for a dependent contribution $c_{\mathrm{dep}}$ was correctly aligned to a precursor contribution $c_{\mathrm{pre}}$, yielding a valid edge $e_j = \langle c_{\mathrm{pre}}, c_{\mathrm{dep}}, a_j \rangle$. This judgment requires the full text of the citing (dependent) paper $d_{\mathrm{dep}}$ to verify that $c_{\mathrm{pre}}$ is a prerequisite of $c_{\mathrm{dep}}$. For example, given the statement \textit{``We built Model X based on Algorithm Y from Smith et al. (2025)''} from paper $d_{\mathrm{dep}}$, we must examine both contributions ($c_{\mathrm{dep}}$: \textit{Model X}, and $c_{\mathrm{pre}}$: \textit{Algorithm Y}) and the full text of the citing paper $d_{\mathrm{dep}}$. We therefore modeled evaluation as an \textsc{LLM-as-a-judge} problem. For 250 randomly sampled edges, we prompted the strongest judge models available at evaluation time (\textsc{GPT-5.5} and \textsc{Sonnet 5}) with $c_{\mathrm{dep}}$, $p_k$, $c_{\mathrm{pre}}$, and $d_{\mathrm{dep}}$, and asked whether the alignment formed a valid prerequisite edge.\footnote{We also piloted a significantly more expensive prompt that included the full text of both the dependent and precursor papers. Reported performance shifted by only $\approx$1\%.}

\begin{table*}
\footnotesize
\centering
\begin{tabular}{lccccccccc}
\toprule
~ & \textbf{Knowledge} & \textbf{Performance} & \textbf{Performance} & \textbf{Perf.} & \textbf{\# Samples} & \textbf{Perf.} & \textbf{Cost} \\
\textbf{Base Model} & \textbf{Cutoff}    & \textbf{(pre-cutoff)} & \textbf{(post-cutoff)} & $\Delta$ & \textbf{(pre/post)} & \textbf{(entire set)} & \textbf{(per 1k)}\\
\midrule
\rowcolor[HTML]{F3F3F3} 
\textsc{Random Baseline}       & --          &  --   &   --  &   --  &   -- / --   &   0.12  &    \cellcolor[HTML]{F3F3F3}\$0   \\
\textsc{IR (tf.idf) Baseline}  & --          &  --   &   --  &   --  &   -- / --   &   0.15  &    \$0   \\
\rowcolor[HTML]{F3F3F3} 
\textsc{Embedding $\dagger$}    & Feb 2024    &  0.10   &   \cellcolor[HTML]{CEE3CE}0.09  &   0.01  &   760 / 1153   &   0.09  &    \cellcolor[HTML]{F3F3F3}\$0.25   \\

\midrule
\textsc{GPT-4o-mini}           & Oct 2023    &  0.16 &   \cellcolor[HTML]{C1E3C1}0.14  &   0.02 &   617 / 1279   &   0.15  & \cellcolor[HTML]{FCF0F0}\$16 \\
\rowcolor[HTML]{F3F3F3} 
\textsc{LLAMA3-70B-InTb}       & Dec 2023    &  0.25 &  \cellcolor[HTML]{A3D5A3}0.21 &   0.04 &   657 / 1250   &  0.23     & \cellcolor[HTML]{FCEDED}\$20 \\
\textsc{GPT-OSS-120B}          & Jun 2024    &  0.25 &   \cellcolor[HTML]{98CD98}0.22  &   0.03  &   845 / 1076  &  0.23  & \cellcolor[HTML]{F1E2E2}\$20 \\
\rowcolor[HTML]{F3F3F3} 
\textsc{GPT-5-nano}            & May 2024    &  0.25 &   \cellcolor[HTML]{9ED39E}0.22  &   0.03  &   817 / 1090  &  0.23  & \cellcolor[HTML]{FEFAFA}\$5  \\
\textsc{QWEN3-235B-A22B}       & Jun 2025    &  0.32 &  \cellcolor[HTML]{7BC47B}0.30 &   0.02 &   1272 / 429   &  0.31     & \cellcolor[HTML]{FDF7F7}\$9 \\
\rowcolor[HTML]{F3F3F3} 
\textsc{GPT-5.4-mini}          & Aug 2025    &  0.37 &   \cellcolor[HTML]{6ABA6A}0.33  &   0.04  &   1364 / 205  &  0.36  & \cellcolor[HTML]{EFDADA}\$29 \\
\textsc{GPT-5-mini}            & May 2024    &  0.42 &   \cellcolor[HTML]{5EB55E}0.36  &   0.06  &   817 / 1090  &  0.38  & \cellcolor[HTML]{F0E0E0}\$22 \\
\rowcolor[HTML]{F3F3F3} 
\textsc{GPT-5.4}               & Aug 2025    &  0.40 &   \cellcolor[HTML]{5CB65C}0.37  &   0.03  &   1364 / 205  &  0.39  & \cellcolor[HTML]{ED9E9E}\$106 \\
\textsc{Claude Haiku 4.5}      & Jun 2025    &  0.41 &   \cellcolor[HTML]{58B458}0.38  &   0.03  &   1337 / 272  &  0.40  & \cellcolor[HTML]{F7D6D6}\$45 \\
\rowcolor[HTML]{F3F3F3} 
\textsc{Claude Sonnet 4.5}     & Jul 2025    &  0.45 &   \cellcolor[HTML]{4DAE4D}0.40  &   0.05  &   1337 / 272  &   0.43 & \cellcolor[HTML]{DF6869}\$160 \\
\textsc{Claude Opus 4.6}       & Aug 2025    &  0.53 &  \cellcolor[HTML]{2CA02C}\textbf{0.48} &   0.05 &   1364 / 205   &  \textbf{0.51}     & \cellcolor[HTML]{D62728}\$235 \\
\bottomrule
\end{tabular}
\vspace{-2mm}
\caption{\footnotesize Model performance on the technological requirement prediction task, reported using temporally-filtered backtesting, where model performance pre- and post- each model's advertised knowledge cutoff is evaluated. Performance is measured using Mean Average Precision (MAP). Cost is reported in USD per 1,000 problems.  Sums of samples (pre/post) may not add to 2000, as problems where only publication years are available (and not specific dates) are discarded when they have the same year as the model's knowledge cutoff to prevent contamination. The Pareto-optimal performance curve is provided in Figure~\ref{fig:pareto-optimal} in the \textsc{Appendix}. $\dagger$~Embedding cosine similarity baseline model is \textsc{OpenAI text-embedding-3-small}.\label{tab:technologicalprecursors}}
\vspace{-2mm}
\end{table*}

Estimated edge alignment error rates are shown in Table~\ref{tab:edge-match-confidence}. Using the same prompt, the judge models applied different decision thresholds: overall \textsc{GPT-5.5} rated 4\% of edges as errors, while \textsc{Sonnet 5} rated 12\% as errors (and 7\% as uncertain). Both agreed that 3\% of edges were incorrect.  When generating edges, the alignment model includes a binary confidence rating (\textit{high, low}), which is provided as  metadata in the graph. When filtering to include only high-confidence edges, overall edge alignment success is estimated at between 92-99\%, while dropping to as low as 83-91\% for low-confidence edges. 

\section{Task: Predicting Technological Requirements}

New scientific advances frequently build upon existing advances: for example, transformers \cite{vaswani-etal-2017-attention} were built upon precursors such as sequence-to-sequence models, attention mechanisms, and word embeddings.  The \textsc{Scientific Contribution Graph} maps such prerequisite relationships, and using these we can build a task to evaluate how well literature-backed automated scientific discovery systems could predict which existing technologies would be needed to develop a new target technology. %

{\flushleft\textbf{Task Framing:}} We operationalize the technological requirement identification task as a ranking task, which mirrors likely deployment scenarios: given a desired new technology, and a list of current technological capabilities, the model must rank candidates such that the most relevant prerequisites for developing the target technology appear highest.  Evaluation is operationalized as temporally filtered backtesting \cite[e.g.][]{jansen-etal-2025-matter,luo-etal-2018-scientific}, where we measure a model's task performance primarily on unseen problems (from papers published after its advertised knowledge cutoff date), which effectively allows evaluating ``new'' technologies unknown to the model.  This form of backtesting is typically used in high-cost domains (such as scientific discovery) where evaluating hundreds or thousands of new discoveries would be prohibitively expensive or infeasible at scale.

{\flushleft\textbf{Data:}} We construct 2,000 technological requirement problems by randomly sampling target scientific contributions from each year in 2021--2025. Each problem consists of a \textit{target technology} and a ranked list of 100 \textit{candidate prerequisite technologies}. Both targets and candidates are scientific contributions from the graph, represented using their \textit{name} and \textit{description} fields. For each sampled target technology, gold prerequisites are drawn from its incoming prerequisite edges in the graph. The remaining candidates are close distractors selected to mirror likely search scenarios: the target name and description are used as an embedding search query over the graph, and the highest-scoring scientific contributions by cosine similarity are returned. Distractors are temporally filtered so that no candidate postdates the target technology; for example, a target from 2024 cannot include distractors from 2025. To reduce false positives, we also remove candidate distractors from papers that contain direct graph edges to other scientific contributions in the same source paper as the target.

{\flushleft\textbf{Models:}} We evaluate a suite of high-performing LLMs using a shared instruction-formatted prompt containing the task description, target technology, and 100 candidate technologies in random order. Each model returns a ranked list in which the most relevant prerequisites appear highest. Because candidates may include both direct prerequisites and more distant related technologies, models are instructed to prefer more immediate precursors; for example, for a \textit{``transformers''} target, \textit{attention mechanisms} should rank above \textit{Markov models}.

{\flushleft\textbf{Results:}} Table~\ref{tab:technologicalprecursors} reports model performance on the technological requirement identification task, using mean average precision (MAP; \citealp{manning2008introduction}), both overall, and split by a dynamic temporal filter (i.e. backtesting) based on whether each problem's source paper was published before or after the model's advertised knowledge cutoff.

The results suggest that recent models are rapidly gaining performance on the technological requirement identification task. While models released approximately two years ago, such as \textsc{GPT-4o-mini}, perform only slightly above a random baseline (0.14 vs.\ 0.12 MAP), the best-performing recent model, \textsc{Claude Opus 4.6}, achieves 0.48 MAP on unseen problems, indicating moderate overall task performance. This performance comes at substantial cost, and the model suite reveals large cost/performance tradeoffs: \textsc{GPT-5-mini} achieves 1\% MAP per \$0.61, while \textsc{Claude Opus 4.6} costs \$4.90 per 1\% MAP. Backtesting reveals models generally perform somewhat better on prediction problems generated from papers published before the model's advertised knowledge cutoff. Performance typically increases by 0.02--0.06 MAP on pre-cutoff problems, suggesting that some overall performance may reflect prior exposure to the relevant literature during model pretraining, rather than purely prospective technological reasoning, though this increase is generally modest across models. %

\section{Discussion}

\begin{figure*}[t!]
    \centering
    \setlength{\fboxsep}{0pt}
    \setlength{\fboxrule}{0.5pt}    
    \fcolorbox{gray}{white}{%
        \clipbox{0pt 1.0cm 0pt 1.0cm}{%
            \includegraphics[
                width=\dimexpr\textwidth-2\fboxrule-2\fboxsep\relax,
                trim=7cm 7cm 7cm 7cm,
                clip
            ]{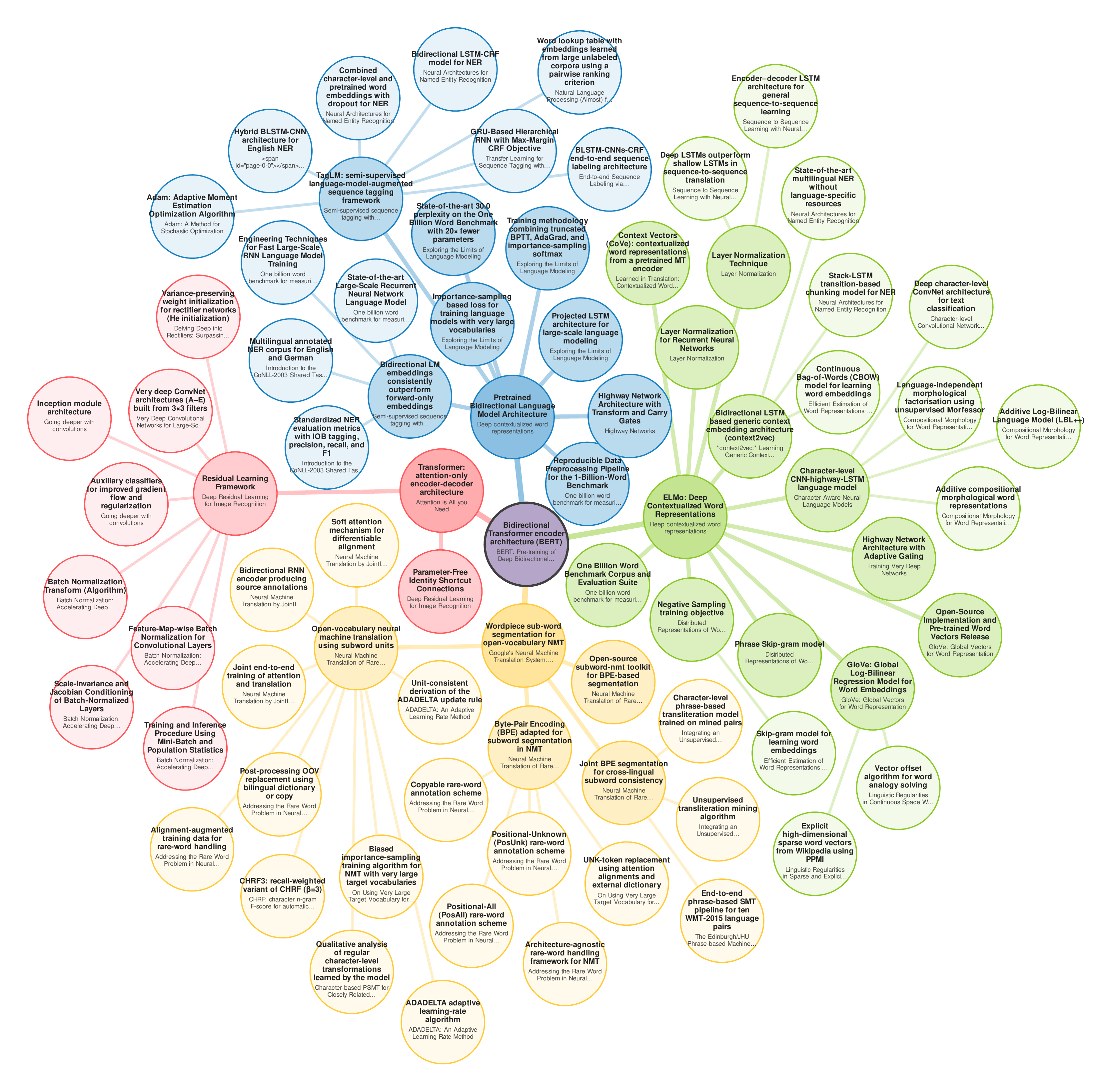}%
        }%
    }    
    \caption{\footnotesize A simplified visualization of the technological roadmap showing precursor technologies that led to the development of the \textit{``Bidirectional Transformer Encoder Architecture (BERT)''} contribution, one of 12 fine-grained contributions extracted from Devlin et al.~\shortcite{devlin-etal-2019-bert}.  Here, nodes represent contributions, while edges represent precursor relations.  All content has been simplified for space: nodes show only contribution names and source paper titles, and not detailed contribution descriptions or metadata; edges do not show precursor descriptions or explanations, and cross-edges between nodes other than the root have been pruned to form a tree for visualization purposes. Precursor relations are shown up to 3 edges away from the root contribution.}
    \label{fig:backwards-graph}
    \vspace{-2mm}
\end{figure*}

As a resource, the \textsc{Scientific Contribution Graph} enables technologies to be traced backward through successive precursor contributions. We expect that the scale and granularity will support downstream investigations challenging to perform with paper-level citation graphs or triple-level graphs alone, including technological roadmapping, impact assessment, and novelty detection.

{\flushleft\textbf{Technological Roadmapping:}}
The \textsc{Scientific Contribution Graph} provides a large-scale technological roadmap, showing how component technologies and findings are combined, extended, and reused in later scientific contributions. This makes it possible to study how technologies emerge from prior work, and visualize local trajectories of scientific development. For example, Figure~\ref{fig:backwards-graph} shows how the Bidirectional Encoder Representations from Transformers (BERT) architecture \cite{devlin-etal-2019-bert} was built from precursor contributions, and how those precursors themselves depend on earlier scientific contributions.

{\flushleft\textbf{Impact Assessment:}}
Assessing the impact of a scientific contribution is challenging and costly, and often approximated using inexpensive but noisy paper-level citation counts -- a citing paper may use a contribution directly, cite it as background, or reflect broader citation biases \cite{balzer2025mitigating,teplitskiy2022status,lariviere2010impact}. 
The \textsc{Scientific Contribution Graph} enables a more direct contribution-level view by tracing downstream contributions that build on it, or new subfields that emerge. For example, \textsc{Appendix~\ref{sec:impact-graphs}} includes large impact assessments of \textit{Retrieval-Augmented Generation (RAG}) \cite{lewis2020retrieval} and \textsc{Self-Instruct Prompting} \cite{wang-etal-2023-self-instruct}, illustrating how these create novel subfields in \textit{internet-augmented prompts} and \textit{fusion-in-decoder QA models} for \textsc{RAG}, or \textit{self-rewarding language models} and \textit{automatic data-poisoning pipelines} for \textsc{Self-Instruct}.
Such analyses may support automated scientific discovery by identifying contribution types that enable high-impact downstream work. They may also help researchers document specific technologies or findings developed from their work, but obscured across citation layers, for example in promotion and tenure cases. This may be through direct textual narratives enabled through these automated impact analyses, or by reporting counts of \textit{new scientific contributions} enabled, directly or indirectly, from one's own scientific contributions.

{\flushleft\textbf{Novelty Detection and Automated Paper Review:}} Automated scientific discovery systems often aim to generate candidate hypotheses that are both novel and underexplored \cite{zheng-etal-2025-automation}. Automated review systems \cite[e.g.][]{afzal-etal-2026-beyond,zhou-etal-2024-llm,Checco2021AIassistedPRA} face a related problem: assessing whether a submission accurately characterizes the novelty of its contributions relative to prior work. Both settings require high-recall search over prior scientific contributions, since a single missed prior contribution may substantially change whether a claim is genuinely novel. By extracting core scientific contributions from full text and making them available for embedding-based search, the \textsc{Scientific Contribution Graph} may provide a scalable intermediate representation for novelty detection and automated review.

{\flushleft\textbf{Scientometrics:}} Computer science papers list an average of 3.2 primary contributions in their introductions \cite{liu2023contributionsum}, whereas our full-text extraction identifies nearly 9 finer-grained contributions per paper, including problem framings, resources, analyses, metrics, and other contribution types. We see this as one example of how the \textsc{Scientific Contribution Graph} may support scientometrics and science-of-science research.

\section{Conclusion}

This work presents the \textsc{Scientific Contribution Graph}, a large-scale resource for technological roadmapping that represents 6 million scientific contributions extracted from 655k source papers, together with 36 million prerequisite relationships that define how each technology builds upon prior contributions. We also formulate technological prerequisite prediction as a downstream evaluation task, and empirically demonstrate that when evaluated using temporally-filtered backtesting, current language models are increasingly able to identify precursor technologies needed to develop new target technologies, suggesting near-term applications in automated scientific discovery. More broadly, we anticipate that the \textsc{Scientific Contribution Graph} will support applications in technological roadmapping, impact assessment, novelty detection, and the study of scientific development. This work is released as open source.

\section*{Limitations}

{\flushleft\textbf{Cost vs Performance:}} The scientific contribution extraction system for knowledge graph construction described in this work has a cost/performance tradeoff that is visible at scale.  While the current \textsc{GPT-OSS-120B} extraction pipeline obtains strong performance, it is possible that higher performance could be obtained by using more expensive models, though this is limited by scaling.  The current scale of the graph (655k source papers) represents approximately 24 months of H200 GPU time ($\approx$\$80,000 in compute costs at current inference provider pricing), though these costs are likely to be mitigated as future open source models return increased performance with lower compute costs.

{\flushleft\textbf{Open Access Articles:}} While the extraction system described in this work can be used on any scientific articles that an organization has access to, the official release includes only open access articles for licensing reasons.  While this is not a significant limitation in the current release domain (natural language processing and artificial intelligence), as much of this domains literature is open access, many high-impact domains such as biology, materials science, and others, have a large proportion of closed-access literature that limit downstream applications including automated scientific discovery.  While this limitation is mitigated by users bringing their own licensed articles, the move towards open-access publishing will help share publicly funded knowledge more freely, and enable further societal benefit by making it easier and more accessible to build further contributions upon those discoveries.

\section*{Acknowledgments}

This research was developed with funding from the Defense Advanced Research Projects Agency’s (DARPA) SciFy program (Agreement No. HR00112520300) to PJ at the University of Arizona. The views expressed are those of the author and do not reflect the official policy or position of the Department of Defense or the U.S. Government. 
PJ has an outside interest in the Allen Institute for Artificial Intelligence. This interest has been disclosed to the University of Arizona and reviewed in accordance with its conflict of interest policies. We thank the three anonymous reviewers for their comments, as well as the members of the DARPA Scientific Feasibility (SciFy) program and the Asta group at the Allen Institute for Artificial Intelligence (\textsc{Ai2}) for thoughtful discussions.

\bibliography{anthology,custom}

\appendix

\section{Contribution Categories}

The list of contribution categories used in the extraction is shown in Table~\ref{tab:contribution_categories}.

\section{Pareto-optimal Performance}
\label{sec:pareto-optimal}

\begin{figure*}[t!]
    \centering
    \includegraphics[scale=0.85]{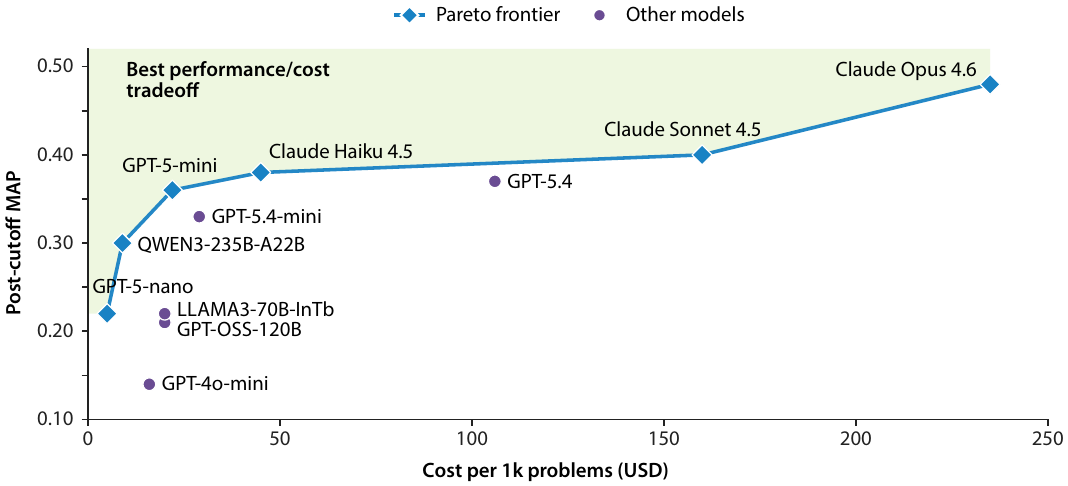}
    \vspace{-3mm}
    \caption{\footnotesize The Pareto frontier for the technological requirement prediction task shown in Table~\ref{tab:technologicalprecursors}, where performance is measured using a model's post-cutoff prediction performance expressed in mean-average precision (MAP), and cost measured as USD per 1,000 problems.  \textsc{OpenAI} models populate the frontier at low cost ranges, while \textsc{Anthropic} models populate the frontier at higher cost ranges.}    
    \label{fig:pareto-optimal}
    \vspace{-2mm}
\end{figure*}

The Pareto frontier for the technological requirement prediction task reported in Table~\ref{tab:technologicalprecursors} is shown in Figure~\ref{fig:pareto-optimal}.

\section{Impact Graphs}
\label{sec:impact-graphs}

\begin{figure*}[t!]
    \centering
    \setlength{\fboxsep}{0pt}
    \setlength{\fboxrule}{0.5pt}    
    \fcolorbox{gray}{white}{%
        \clipbox{0pt 0cm 0pt 0cm}{%
            \includegraphics[
                width=\dimexpr\textwidth-2\fboxrule-2\fboxsep\relax,
                trim=20cm 20cm 20cm 20cm,
                clip
            ]{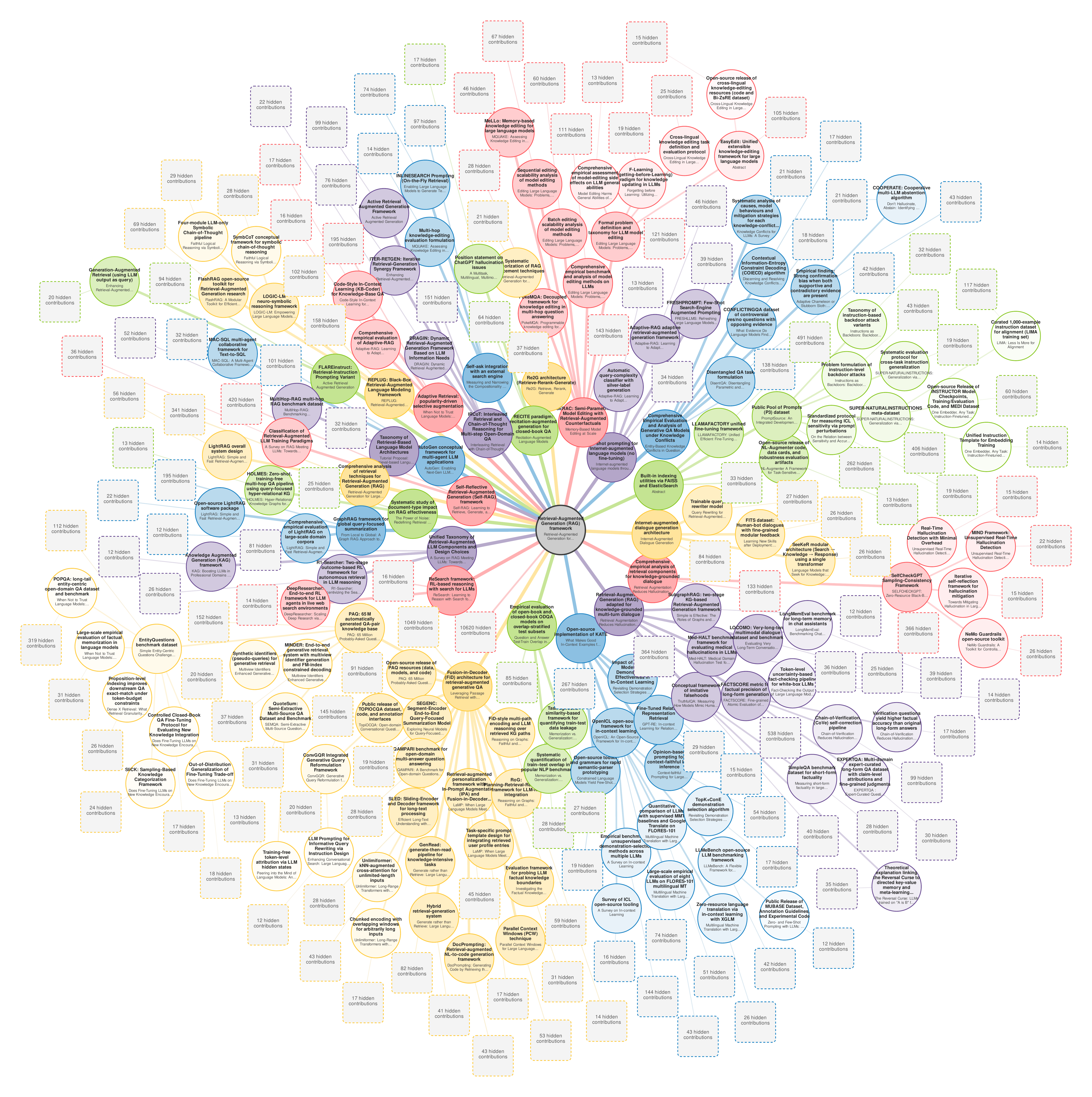}%
        }%
    }    
    \caption{\footnotesize A simplified impact assessment for the \textit{``Retrieval Augmented Genration (RAG) Framework'' contribution} from \cite{lewis2020retrieval}.  Here, nodes represent contributions, while edges represent precursor relations \textit{in the opposite direction} -- i.e. that the displayed technologies were build \textit{upon} the root node.  All content has been simplified for space: nodes show only contribution names and source paper titles, and not detailed contribution descriptions or metadata; edges do not show precursor descriptions or explanations, and cross-edges between nodes other than the root have been pruned to form a tree for visualization purposes. Similarly, only the highest linked contributions are shown, while total numbers of hidden contributions are shown in the square nodes.}
    \label{fig:impact-graph1}
    \vspace{-2mm}
\end{figure*}

\begin{figure*}[t!]
    \centering
    \setlength{\fboxsep}{0pt}
    \setlength{\fboxrule}{0.5pt}    
    \fcolorbox{gray}{white}{%
        \clipbox{0pt 0cm 0pt 0cm}{%
            \includegraphics[
                width=\dimexpr\textwidth-2\fboxrule-2\fboxsep\relax,
                trim=8cm 8cm 8cm 8cm,
                clip
            ]{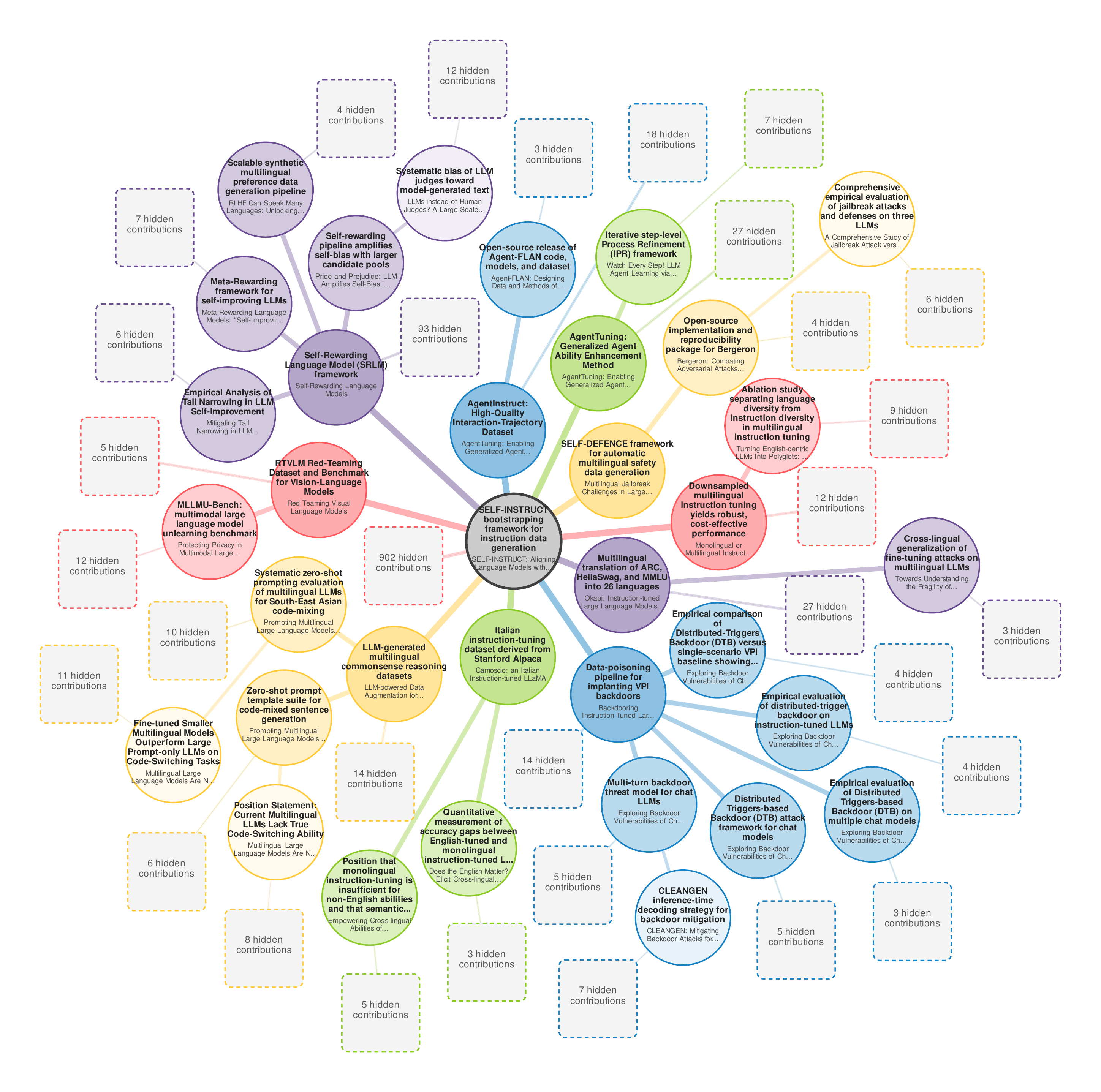}%
        }%
    }    
    \caption{\footnotesize A simplified impact assessment for the \textit{``\textsc{Self-Instruct} bootstrapping framework for instruction data generation'' contribution} from \cite{wang-etal-2023-self-instruct}.  Here, nodes represent contributions, while edges represent precursor relations \textit{in the opposite direction} -- i.e. that the displayed technologies were build \textit{upon} the root node.  All content has been simplified for space: nodes show only contribution names and source paper titles, and not detailed contribution descriptions or metadata; edges do not show precursor descriptions or explanations, and cross-edges between nodes other than the root have been pruned to form a tree for visualization purposes. Similarly, only the highest linked contributions are shown, while total numbers of hidden contributions are shown in the square nodes.}
    \label{fig:impact-graph2}
    \vspace{-2mm}
\end{figure*}

{\flushleft\textbf{Impact Graphs (single contribution):}} Example impact assessment graphs are shown for \textit{Retrieval Augmented Generation (RAG)} in Figure~\ref{fig:impact-graph1}, and \textit{Self-Instruct Bootstrapping} in Figure~\ref{fig:impact-graph2}.

{\flushleft\textbf{Impact Tables (whole paper):}} Numerical impact assessments analogous to citation counts can be obtained by counting the number of downstream contributions (out to some depth) that depend upon a given contribution.  This functionality is also provided in the \textsc{API}.  Examples of impact assessments for several papers are shown in Tables~\ref{tab:impact_rag}, ~\ref{tab:impact_selfinstruct}, ~\ref{tab:impact_entailmentbank}.

\section{Prompts}

Example prompts are provided in listings below.

\section{Example Contribution Record}
\label{sec:example-contributions}
An example \textsc{JSON} extraction record for a single extracted contribution (the \textit{Bidirectional Transformer Encoder Architecture; BERT}) of 12 total contributions extracted from Devlin et al.~\shortcite{devlin-etal-2019-bert} is shown in Listings~\ref{fig:json-a} through~\ref{fig:json-c}.

\section{AI Usage Disclosure}
AI tools were used in the preparation of this work, centrally in the following ways: (1) Condensing and polishing a human-written draft manuscript, (2) \textsc{GitHub Copilot} was used to suggest completions when writing primarily human-authored code, (3) The following code components were primarily \textsc{LLM} authored: the embedding search, and the web demo. 

\newpage

\begin{table*}[t!]
\footnotesize
\centering
\begin{tabular}{lp{0.6\linewidth}}
\toprule
\textbf{Category} & \textbf{Description} \\
\midrule
\rowcolor[HTML]{F3F3F3} 
\texttt{problem\_formulation} & Reconceiving a task in a novel way, or identifying new angles or perspectives on long-standing problems. Can also be used to propose a new kind of task. \\
\texttt{theoretical\_insight} & Proposing new theories, either from deriving theoretical results or directly from empirical or analysis results. \\
\rowcolor[HTML]{F3F3F3} 
\texttt{conceptual\_framework} & Developing high-level conceptual frameworks that help guide future research. \\
\texttt{resource\_benchmark} & Creating a new benchmark that evaluates model performance on a task. \\
\rowcolor[HTML]{F3F3F3} 
\texttt{resource\_dataset} & Creating new datasets that are primarily generic resources, without the intention of being used as a benchmark (i.e., without training/evaluation sets). \\
\texttt{tool\_system\_software} & Software, tools, systems, code, or other artifacts whose primary purpose is to be used by other researchers. \\
\rowcolor[HTML]{F3F3F3} 
\texttt{empirical\_evaluation} & Presenting novel empirical results discovered through experimentation, such as by running a model on a benchmark. \\
\texttt{analysis} & Presenting novel analyses of data that yield new insights. \\
\rowcolor[HTML]{F3F3F3} 
\texttt{models\_or\_architectures} & Proposing completely novel model designs, or modifications to existing models or architectures. \\
\texttt{techniques\_algorithms} & Introducing new learning algorithms, optimization techniques, or other algorithmic contributions. \\
\rowcolor[HTML]{F3F3F3} 
\texttt{representational} & Novel ways to represent or encode data. \\
\texttt{research\_methods\_procedures} & Establishing novel experimental protocols, methods, or procedures for conducting aspects of research. \\
\rowcolor[HTML]{F3F3F3} 
\texttt{metrics\_instruments} & Developing novel evaluation metrics or protocols for assessment. \\
\texttt{position\_statement} & Articulating a clear position on a debated topic, supported by evidence or argumentation. \\
\rowcolor[HTML]{F3F3F3} 
\texttt{real\_world\_application} & Demonstrations that a research idea works in a practical, real-world setting. \\
\texttt{society\_ethics\_policy} & Addressing ethical, societal, and/or policy-related issues. \\
\rowcolor[HTML]{F3F3F3} 
\texttt{other} & Any other type of scientific contribution or claim not covered here. \\
\bottomrule
\end{tabular}
\caption{\footnotesize While scientific contribution extraction categories are  unrestricted, the above categories of scientific contributions are provided in the extraction schema as examples to ground the fine-grained granularity of the contribution extraction step.\label{tab:contribution_categories}}
\end{table*}

\begin{table*}[t!]
  \footnotesize
  \centering
  \begin{tabular}{lcc@{\hspace{2.5em}}lcc}
  \toprule
  \textbf{Field of Study} & \textbf{Papers} & \textbf{\%} & \textbf{Field of Study} & \textbf{Papers} & \textbf{\%} \\
  \midrule
  \rowcolor[HTML]{F3F3F3}
  Medicine & 304,898 & 46.5\% & Sociology & 4,796 & 0.7\% \\
  Computer Science & 272,796 & 41.6\% & Education & 3,637 & 0.6\% \\
  \rowcolor[HTML]{F3F3F3}
  Biology & 177,090 & 27.0\% & Political Science & 3,108 & 0.5\% \\
  Physics & 163,066 & 24.9\% & Business & 2,642 & 0.4\% \\
  \rowcolor[HTML]{F3F3F3}
  Mathematics & 58,616 & 8.9\% & Economics & 2,628 & 0.4\% \\
  Linguistics & 51,830 & 7.9\% & Agricultural and Food Sciences & 2,615 & 0.4\% \\
  \rowcolor[HTML]{F3F3F3}
  Engineering & 51,458 & 7.9\% & Geography & 1,904 & 0.3\% \\
  Environmental Science & 40,636 & 6.2\% & Philosophy & 1,900 & 0.3\% \\
  \rowcolor[HTML]{F3F3F3}
  Chemistry & 35,962 & 5.5\% & Law & 1,650 & 0.3\% \\
  Materials Science & 20,198 & 3.1\% & History & 1,111 & 0.2\% \\
  \rowcolor[HTML]{F3F3F3}
  Psychology & 13,494 & 2.1\% &  &  &  \\
  \bottomrule
  \end{tabular}
  \caption{\footnotesize Semantic Scholar fields of study for the 655k papers in the scientific contribution graph, for fields with more than 1,000 papers. Papers may be tagged with more than one field (1.9 on average), so percentages sum to more than 100\%.\label{tab:s2_fields_of_study}}
\end{table*}

\begin{table*}[t!]
  \centering
  \footnotesize
  \setlength{\tabcolsep}{4pt}
  \begin{tabular}{p{0.30\linewidth}p{0.20\linewidth}cccc}
  \toprule
   & & \multicolumn{2}{c}{\textbf{Annotator 1}} & \multicolumn{2}{c}{\textbf{Annotator 2}} \\
  \cmidrule(lr){3-4}\cmidrule(lr){5-6}
  \textbf{Alignment Category} & \textbf{Symbol} & \textbf{Count} & \textbf{Prop.} & \textbf{Count} & \textbf{Prop.} \\
  \midrule
  \rowcolor{posA}
  Direct Matches & $R \rightarrow M$ & 99 & 38\% & 78 & 29\% \\
  \rowcolor{posA}
  Additional Valid Contributions & $M \rightarrow \text{unmatched}$ & 64 & 25\% & 89 & 33\% \\
  \rowcolor{posA}
  High-Resolution Decomp. & $R \rightarrow (M_1,\ldots,M_n)$ & 50 & 19\% & 71 & 27\% \\
  \rowcolor{warnA}
  Coarser Decomposition & $(R_1,\ldots,R_n) \rightarrow M$ & 23 & 9\% & 14 & 5\% \\
  \rowcolor{negA}
  Missed Contributions & $R \rightarrow \text{unmatched}$ & 22 & 9\% & 15 & 6\% \\
  \midrule
  \rowcolor[HTML]{F3F3F3}
  Total & & 258 & 100\% & 267 & 100\% \\
  \bottomrule
  \end{tabular}
  \vspace{-2mm}
  \caption{\footnotesize Alignment between reference-set (R) and model-based (M) scientific contribution extraction for two independent annotators, and companion study to Table~\ref{tab:contribution-alignment}. Annotator 1 labeled 25 papers and Annotator 2 labeled 26 papers, with the sets overlapping on 9 papers.  Overall, the results from both annotators show an overall high extraction quality -- the extraction model is generally picking up scientific contributions that are missed by human domain experts, or decomposing those contributions at a higher-resolution than the human experts, while missing only a small number of contributions labeled by the human experts.}
  \label{tab:contribution-alignment-both}
  \vspace{-2mm}
\end{table*}

\begin{table*}[t!]
  \footnotesize
  \centering
  \begin{tabular}{lp{0.52\linewidth}cc}
  \toprule
  \multicolumn{4}{l}{\textbf{Impact Assessment:} Retrieval-Augmented Generation for Knowledge-Intensive NLP Tasks \cite{lewis2020retrieval}} \\
  \midrule
  \textbf{Contribution ID} & \textbf{Contribution} & \textbf{Impact} & \textbf{Impact (damp.)} \\
  \midrule
  \rowcolor[HTML]{F3F3F3} 
  \texttt{218869575.c0} & Retrieval-Augmented Generation (RAG) framework & 33,038 & 14,045.1 \\
  \texttt{218869575.c1-2} & RAG-Token variant & 25,298 & 9,490.5 \\
  \rowcolor[HTML]{F3F3F3} 
  \texttt{218869575.c1-1} & RAG-Sequence variant & 25,265 & 9,467.0 \\ 
  \texttt{218869575.c8} & Hot-swapping of the non-parametric knowledge index & 16,952 & 4,573.1 \\
  \rowcolor[HTML]{F3F3F3} 
  \texttt{218869575.c2} & End-to-end fine-tuning without document-level supervision & 12,408 & 4,641.9 \\
  \texttt{218869575.c3} & Decoding strategies for RAG-Sequence (Thorough vs Fast) & 7,458 & 2,518.6 \\
  \rowcolor[HTML]{F3F3F3} 
  \texttt{218869575.c5} & Fact verification with RAG via sequence-label formulation (FEVER) & 7,185 & 2,316.5 \\
  \texttt{218869575.c6} & Ablation analysis of retrieval strategies and retrieved document count & 7,028 & 2,159.1 \\
  \rowcolor[HTML]{F3F3F3} 
  \texttt{218869575.c9} & Open-source release of RAG code, pretrained models, and index pipeline & 5,444 & 1,748.9 \\
  \texttt{218869575.c4-1} & State-of-the-art Exact Match on four open-domain QA benchmarks & 4,194 & 1,152.3 \\
  \rowcolor[HTML]{F3F3F3} 
  \texttt{218869575.c7} & Analysis of generation diversity, factuality, specificity, and controllable factual content in Jeopardy question generation & 2,437 & 704.6 \\
  \texttt{218869575.c4-4} & Near-state-of-the-art fact verification on FEVER & 2,130 & 616.0 \\
  \rowcolor[HTML]{F3F3F3} 
  \texttt{218869575.c4-2} & State-of-the-art abstractive QA performance on MS-MARCO & 2,095 & 594.4 \\
  \texttt{218869575.c4-3} & State-of-the-art Jeopardy-style question generation & 2,078 & 585.4 \\
  \midrule
  \multicolumn{2}{l}{\textbf{Paper-level total} (unique downstream contributions)} & \textbf{33,358} & \textbf{14,610.7} \\
  \bottomrule
  \end{tabular}
  \caption{\footnotesize Example impact assessment for each scientific contribution extracted from \textit{Retrieval-Augmented Generation for Knowledge-Intensive NLP Tasks} \cite{lewis2020retrieval}, computed by forward-crawling the contribution graph to a maximum depth of 5. \textbf{Impact} counts the unique downstream scientific contributions that build upon a given contribution, directly or indirectly. \textbf{Impact (dampened)} applies reciprocal-depth weighting, so that contributions at depth 1 count 1, depth 2 count $\frac{1}{2}$, depth 3 count $\frac{1}{3}$, and so on. The paper-level total counts unique downstream contributions across all contributions in the paper, and is therefore smaller than the column sum.\label{tab:impact_rag}}
\end{table*}

\begin{table*}[t!]
    \footnotesize
    \centering
    \begin{tabular}{lp{0.52\linewidth}cc}
    \toprule
    \multicolumn{4}{l}{\textbf{Impact Assessment:} SELF-INSTRUCT: Aligning Language Models with Self-Generated Instructions \cite{wang-etal-2023-self-instruct}} \\
    \midrule
    \textbf{Contribution ID} & \textbf{Contribution} & \textbf{Impact} & \textbf{Impact (damp.)} \\
    \midrule
    \rowcolor[HTML]{F3F3F3}
    \texttt{254877310.c0} & SELF-INSTRUCT bootstrapping framework for instruction data generation & 2,858 & 1,536.4 \\
    \texttt{254877310.c4} & Large synthetic instruction dataset (\textasciitilde{}52 K instructions, 82 K instances) & 481 & 288.3 \\
    \rowcolor[HTML]{F3F3F3}
    \texttt{254877310.c5} & Empirical improvement on SUPER-NATURALINSTRUCTIONS benchmark via SELF-INSTRUCT fine-tuning & 451 & 255.7 \\
    \texttt{254877310.c2} & Few-shot classification-task identification & 148 & 76.0 \\
    \rowcolor[HTML]{F3F3F3}
    \texttt{254877310.c9-2} & Open-source release of SELF-INSTRUCT synthetic instruction dataset & 136 & 79.3 \\
    \texttt{254877310.c1} & Dual instance-generation strategies (input-first vs. output-first) for classification tasks & 102 & 43.3 \\
    \rowcolor[HTML]{F3F3F3}
    \texttt{254877310.c3-1} & ROUGE-L similarity-based instruction deduplication filter & 91 & 43.0 \\
    \texttt{254877310.c3-4} & Duplicate input/output detection filter & 78 & 32.6 \\
    \rowcolor[HTML]{F3F3F3}
    \texttt{254877310.c8-2} & Data quality improvement via output regeneration with InstructGPT-003 & 55 & 26.3 \\
    \texttt{254877310.c6-1} & Curated benchmark of 252 expert-written user-oriented instructions & 21 & 16.0 \\
    \rowcolor[HTML]{F3F3F3}
    \texttt{254877310.c8-1} & Data size scaling analysis for SELF-INSTRUCT & 20 & 13.3 \\
    \texttt{254877310.c9-1} & Open-source release of SELF-INSTRUCT codebase & 8 & 7.5 \\
    \rowcolor[HTML]{F3F3F3}
    \texttt{254877310.c3-3} & Length-based instruction and instance sanity filter & 7 & 7.0 \\
    \texttt{254877310.c6-2} & Human evaluation of GPT3SELF-INST on the 252-instruction benchmark & 7 & 4.7 \\
    \rowcolor[HTML]{F3F3F3}
    \texttt{254877310.c7-2} & Quality assessment of generated instructions & 6 & 6.0 \\
    \texttt{254877310.c3-2} & Keyword-based modality exclusion filter & 4 & 4.0 \\
    \rowcolor[HTML]{F3F3F3}
    \texttt{254877310.c7-1} & Diversity analysis of generated instructions & 4 & 4.0 \\
    \midrule
    \multicolumn{2}{l}{\textbf{Paper-level total} (unique downstream contributions)} & \textbf{3,142} & \textbf{1,724.9} \\
    \bottomrule
    \end{tabular}
    \caption{\footnotesize Example impact assessment for each scientific contribution extracted from \textit{SELF-INSTRUCT: Aligning Language Models with Self-Generated Instructions} \cite{wang-etal-2023-self-instruct}, computed by
  forward-crawling the contribution graph to a maximum depth of 5. \textbf{Impact} counts the unique downstream scientific contributions that build upon a given contribution, directly or indirectly. \textbf{Impact
  (dampened)} applies reciprocal-depth weighting, so that contributions at depth 1 count 1, depth 2 count $\frac{1}{2}$, depth 3 count $\frac{1}{3}$, and so on. The paper-level total counts unique downstream
  contributions across all contributions in the paper, and is therefore smaller than the column sum.\label{tab:impact_selfinstruct}}
\end{table*}

  \begin{table*}[t!]
    \footnotesize
    \centering
    \begin{tabular}{lp{0.52\linewidth}cc}
    \toprule
    \multicolumn{4}{l}{\textbf{Impact Assessment:} Explaining Answers with Entailment Trees \cite{dalvi-etal-2021-explaining}} \\
    \midrule
    \textbf{Contribution ID} & \textbf{Contribution} & \textbf{Impact} & \textbf{Impact (damp.)} \\
    \midrule
    \rowcolor[HTML]{F3F3F3}
    \texttt{233297051.c1} & ENTAILMENTBANK: a large-scale dataset of multistep entailment trees for elementary-science QA & 774 & 390.4 \\
    \texttt{233297051.c4} & EntailmentWriter: T5-based generative model for entailment-tree generation & 96 & 56.3 \\
    \rowcolor[HTML]{F3F3F3} 
    \texttt{233297051.c0} & Formulation of explanation as multistep entailment trees & 82 & 50.3 \\
    \texttt{233297051.c5} & Tree Alignment Algorithm for Evaluating Entailment Trees & 30 & 28.5 \\
    \rowcolor[HTML]{F3F3F3} 
    \texttt{233297051.c3} & Multi-stage relevant-fact retrieval pipeline for the full-corpus explanation task & 19 & 9.0 \\
    \texttt{233297051.c7} & Analysis of reasoning types required for multistep entailments & 17 & 12.5 \\
    \rowcolor[HTML]{F3F3F3} 
    \texttt{233297051.c8-2} & Shredded-tree interactive explanation generation & 14 & 9.0 \\
    \texttt{233297051.c2} & Web-based drag-and-drop authoring tool for constructing entailment trees & 3 & 3.0 \\
    \rowcolor[HTML]{F3F3F3} 
    \texttt{233297051.c6-2} & Error analysis of EntailmentWriter revealing common failure modes & 3 & 3.0 \\
    \texttt{233297051.c6-1} & Empirical evaluation of EntailmentWriter on three explanation tasks & 0 & 0.0 \\
    \rowcolor[HTML]{F3F3F3} 
    \texttt{233297051.c8-1} & Zero-shot out-of-domain generalization of EntailmentWriter & 0 & 0.0 \\
    \midrule
    \multicolumn{2}{l}{\textbf{Paper-level total} (unique downstream contributions)} & \textbf{790} & \textbf{416.9} \\
    \bottomrule
    \end{tabular}
    \caption{\footnotesize Example impact assessment for each scientific contribution extracted from \textit{Explaining Answers with Entailment Trees} \cite{dalvi-etal-2021-explaining}, computed by forward-crawling the contribution graph
  to a maximum depth of 5. \textbf{Impact} counts the unique downstream scientific contributions that build upon a given contribution, directly or indirectly. \textbf{Impact (dampened)} applies reciprocal-depth
  weighting, so that contributions at depth 1 count 1, depth 2 count $\frac{1}{2}$, depth 3 count $\frac{1}{3}$, and so on. The paper-level total counts unique downstream contributions across all contributions in the
  paper, and is therefore smaller than the column sum.\label{tab:impact_entailmentbank}}
\end{table*}

\begin{figure*}[t]
\begin{lstlisting}[language=json, caption={Example detailed \textsc{JSON} extraction record for one of 12 contributions from Devlin et al.~\shortcite{devlin-etal-2019-bert}, showing detailed node and edge content, including descriptions, explanations, and contribution alignments. Across all 12 contributions, prerequisties, and alignments, the full extraction record for this paper contains 26,497 tokens \textit{(part 1 of 3)}.},label={fig:json-a}]
{
    "corpus_id": "52967399",
    "title": "BERT: Pre-training of Deep Bidirectional Transformers for Language Understanding",
    "year": 2019,
    "contributions": [
        {
            "contribution_id": "52967399.c0",
            "name": "Bidirectional Transformer encoder architecture (BERT)",
            "description": "BERT introduces a multi-layer Transformer encoder in which every token attends to both its left and right context through fully bidirectional self-attention. The model combines WordPiece token embeddings, learned segment embeddings, positional embeddings, and special tokens ([CLS], [SEP]) to represent single sentences or sentence pairs uniformly. This architecture serves as a universal language representation that can be fine-tuned with a minimal task-specific output layer for a wide range of NLP tasks.",
            "types": [
                {
                    "type": "models_or_architectures",
                    "explanation": "Proposes a new model architecture based on a bidirectional Transformer encoder."
                },
                {
                    "type": "representational",
                    "explanation": "Defines a novel way to encode text that captures full left-right context at every layer."
                }
            ],
            "sections": [
                "Section 3: BERT",
                "Section 3.1: Pre-training BERT",
                "Appendix A.4: Comparison of BERT, ELMo, and OpenAI GPT"
            ],
            "prerequisites": [
                {
                    "name": "Transformer encoder (self-attention) architecture",
                    "description": "The Transformer encoder introduced by Vaswani et al. (2017) uses multi-head self-attention, positional embeddings, and feed-forward sub-layers to process sequences in parallel, enabling deep contextual representations.",
                    "explanation": "BERT directly builds on the Transformer encoder, replacing the unidirectional decoder with a fully bidirectional encoder.",
                    "core_or_peripheral": "core",
                    "references": [
                        {
                            "type": "paper",
                            "paper_title": "Attention is all you need",
                            "paper_year": 2017,
                            "paper_venue": "Advances in Neural Information Processing Systems",
                            "corpus_id": "13756489",
                            "matches": [
                                {
                                    "contribution_id": "13756489.c0",
                                    "explanation": "Describes the full Transformer encoder-decoder architecture, including multi-head self-attention, positional embeddings, and feed-forward sub-layers that constitute the encoder BERT builds on.",
                                    "match_type": "strong",                                    
                                },
                                {
                                    "contribution_id": "13756489.c1",
                                    "explanation": "Introduces Scaled Dot-Product Attention, the core operation used in the encoder's self-attention layers.",
                                    "match_type": "weak",                                    
                                },
                                {
                                    "contribution_id": "13756489.c2",
                                    "explanation": "Presents Multi-Head Attention, a key component that enables the encoder to attend to multiple representation sub-spaces in parallel.",
                                    "match_type": "weak",                                    
                                },
                                {
                                    "contribution_id": "13756489.c3",
                                    "explanation": "Provides Sinusoidal Positional Encodings, the deterministic positional embedding method employed by the encoder.",
                                    "match_type": "weak",                                    
                                }
                            ]
                        }
                    ]
                },

                ... (continued)
\end{lstlisting}
\end{figure*}

\lstset{firstnumber=last}   %
\begin{figure*}[t]
\begin{lstlisting}[language=json, caption={Example detailed \textsc{JSON} extraction record for one of 12 contributions from Devlin et al.~\shortcite{devlin-etal-2019-bert}, showing detailed node and edge content, including descriptions, explanations, and contribution alignments. Across all 12 contributions, prerequisties, and alignments, the full extraction record for this paper contains 26,497 tokens \textit{(part 2 of 3)}.},label={fig:json-b}]
                ... (continued)
                
                {
                    "name": "WordPiece subword tokenization",
                    "description": "WordPiece splits words into subword units based on a learned vocabulary, allowing open-vocabulary handling and efficient embedding lookup.",
                    "explanation": "BERT's input representation sums token, segment, and position embeddings derived from WordPiece tokens.",
                    "core_or_peripheral": "core",
                    "references": [
                        {
                            "type": "paper",
                            "paper_title": "Google's neural machine translation system: Bridging the gap between human and machine translation",
                            "paper_year": 2016,
                            "paper_venue": "arXiv preprint",
                            "corpus_id": "3603249",
                            "matches": [
                                {
                                    "contribution_id": "3603249.c3",
                                    "explanation": "The cited contribution introduces the WordPiece subword segmentation model, which is the same tokenization method required by BERT.",
                                    "match_type": "strong",
                                }
                            ]
                        }
                    ]
                },
                {
                    "name": "Understanding limitations of unidirectional language models",
                    "description": "Prior work (e.g., OpenAI GPT, ELMo) showed that left-to-right or right-to-left language models cannot condition on both sides simultaneously, limiting performance on tasks that require full context.",
                    "explanation": "Motivates the need for a bidirectional pre-training architecture, which BERT provides.",
                    "core_or_peripheral": "core",
                    "references": [
                        {
                            "type": "paper",
                            "paper_title": "Improving language understanding with unsupervised learning",
                            "paper_year": 2018,
                            "paper_venue": "Technical report, OpenAI",
                            "corpus_id": null # Paper not yet added/processed, or full-text unavailable in open corpus
                            "matches": []
                        },
                        {
                            "type": "paper",
                            "paper_title": "Deep contextualized word representations",
                            "paper_year": 2018,
                            "paper_venue": "NAACL",
                            "corpus_id": "3626819",
                            "matches": [
                                {
                                    "contribution_id": "3626819.c0",
                                    "explanation": "ELMo introduces deep contextualized word representations derived from a bidirectional LSTM language model, directly demonstrating that a bidirectional architecture can overcome the context-conditioning limits of unidirectional models.",
                                    "match_type": "strong",                                    
                                },
                                {
                                    "contribution_id": "3626819.c1",
                                    "explanation": "The paper details the design of a pretrained bidirectional LSTM language model, providing the concrete architectural foundation that addresses the inability of left-to-right or right-to-left models to condition on both sides.",
                                    "match_type": "strong",                                    
                                }
                            ]
                        }
                    ]
                },

                ... (continued)
\end{lstlisting}
\end{figure*}

\lstset{firstnumber=last}   %
\begin{figure*}[t]
\begin{lstlisting}[language=json, caption={Example detailed \textsc{JSON} extraction record for one of 12 contributions from Devlin et al.~\shortcite{devlin-etal-2019-bert}, showing detailed node and edge content, including descriptions, explanations, and contribution alignments. Across all 12 contributions, prerequisties, and alignments, the full extraction record for this paper contains 26,497 tokens \textit{(part 3 of 3)}.},label={fig:json-c}]
                ... (continued)
                
                {
                    "name": "Masked Language Model (MLM) pre-training objective",
                    "description": "MLM randomly masks a subset of tokens and trains the model to predict the original token IDs using both left and right context, enabling deep bidirectional learning.",
                    "explanation": "BERT's bidirectional encoder is trained with this objective; without it the model would not learn to fuse full context.",
                    "core_or_peripheral": "peripheral",
                    "references": [
                        {
                            "type": "internal",
                            "contribution_name": "Masked Language Model (MLM) pre-training objective",
                            "contribution_id": "52967399.c1",
                            "explanation": "MLM provides the learning signal that makes the bidirectional Transformer effective."
                        }
                    ]
                },
                {
                    "name": "Next Sentence Prediction (NSP) pre-training task",
                    "description": "NSP trains the model to predict whether two sentences are consecutive in the original corpus, encouraging the encoder to capture inter-sentence relationships.",
                    "explanation": "NSP supplies sentence-pair level supervision that complements MLM and is used during BERT pre-training.",
                    "core_or_peripheral": "peripheral",
                    "references": [
                        {
                            "type": "internal",
                            "contribution_name": "Next Sentence Prediction (NSP) pre-training task",
                            "contribution_id": "52967399.c2",
                            "explanation": "NSP is part of the pre-training pipeline that shapes the representations learned by the bidirectional encoder."
                        }
                    ]
                },
                {
                    "name": "Large-scale unsupervised pre-training data and compute",
                    "description": "BERT is trained on the BooksCorpus (~800M words) and English Wikipedia (~2.5B words) with large batch sizes on Cloud TPUs, providing the data volume and compute needed for deep models.",
                    "explanation": "The depth and capacity of the bidirectional Transformer require massive data and compute to avoid over-fitting and to learn general language knowledge.",
                    "core_or_peripheral": "peripheral",
                    "references": [
                        {
                            "type": "internal",
                            "contribution_name": "Large-scale pre-training methodology (data, batch size, curriculum)",
                            "contribution_id": "52967399.c4",
                            "explanation": "Describes the corpus, batch size, and training schedule that enable successful training of the bidirectional encoder."
                        }
                    ]
                },
                {
                    "name": "Optimization techniques (Adam optimizer, learning-rate warm-up, GELU activation)",
                    "description": "Training uses the Adam optimizer with weight decay, a warm-up schedule for the learning rate, and the Gaussian Error Linear Unit (GELU) nonlinearity, which improve convergence of deep Transformers.",
                    "explanation": "These techniques are required to reliably train the deep bidirectional Transformer without instability.",
                    "core_or_peripheral": "peripheral",
                    "references": [
                        {
                            "type": "paper",
                            "paper_title": "Bridging nonlinearities and stochastic regularizers with Gaussian error linear units",
                            "paper_year": 2016,
                            "paper_venue": "CoRR",
                            "corpus_id": "2359786",
                            "matches": []  # No alignments to other contributions for this specific prerequisite. This is an example of an edge before alignment, which typically happens when paper full-text for the referenced article is unavailable in the open research corpus.
                        }
                    ]
                }
            ]
        },
        ... (truncated; other contributions from this paper not shown here for space)
\end{lstlisting}
\end{figure*}

\lstset{firstnumber=1}   %
\begin{figure*}[t]
\begin{lstlisting}[language=json, caption={Example prompt for the contribution extraction subtask \textit{(part 1 of 2)}.}]
# Contribution Extraction Prompt 

You are an incredibly capable scientific research assistant. You always follow instructions carefully and accurately, and never make up information. You proceed with the highest scientific rigor and integrity.

Broad Task Context: Your broad task is in technological roadmapping: identifying the scientific contributions of research papers, and what prerequisite knowledge or capabilities were required to develop those scientific contributions.
The ultimate goal is to build a graph of scientific contributions, that clearly identifies how previous scientific contributions led to later scientific contributions, to better understand how making contributions today could lead to different kinds of capabilities in the future.

The task instructions come AFTER the paper. Treat them as the highest priority.

NOTE: The paper text was automatically extracted from the PDF, and may contain some extraction errors.
=== PAPER START ===
```text
<<VARIABLE: paper_text>>
```
=== PAPER END ===

=== TASK START ===

# Broad Task Context
Your broad task is in technological roadmapping: identifying the scientific contributions of research papers, and what prerequisite knowledge or capabilities were required to develop those scientific contributions.
The ultimate goal is to build a graph of scientific contributions, that clearly identifies how previous scientific contributions led to later scientific contributions, to better understand how making contributions today could lead to different kinds of capabilities in the future.

# Task
Your specific task is to read a research paper, and extract its scientific contributions/claims, and identify what prerequisite knowledge was required to make these contributions (nominally, with references to that prerequisite knowledge, if they are listed in the paper).
This will be a multi-step process:
Step 1: Read the paper carefully and identify its scientific contributions/claims
Step 2: For each contribution/claim, identify the prerequisite knowledge or capabilities required to make that contribution/claim
Step 3: For each prerequisite knowledge or capability, identify references to that prerequisite knowledge in the paper (if any)

Currently we are on *Step 1: Read the paper carefully and identify its scientific contributions/claims*

# Contribution/Claim Types
Here are some (non-exhaustive!) high-level types of scientific contributions/claims that are often made in research papers.  These may help guide your discovery/framing of scientific contributions/claims. Remember, this list is non-exhaustive, and other types of contributions/claims may exist.
When identifying contributions/claims, please classify each contribution/claim into one or more of these types (or `other` if none fit well):
1. `problem_formulation`: Reconceiving a task in a novel way, or identifying new angles/perspectives on long-standing problems. Can also be used to propose a new kind of task.
2. `theoretical_insight`: Proposing new theories, either from deriving theoretical results, or directly from empirical/analysis results.
3. `conceptual_framework`: Developing high-level conceptual frameworks that help guide future research.
4. `resource_benchmark`: Creating a new benchmark that evaluates model performance on a task.
5. `resource_dataset`: Creating new datasets that are primarily generic resources, without the intention of being used as a benchmark (i.e. without training/evaluation sets).
6. `tool_system_software`: Software, tools, systems, code, or other artifacts whose primary purpose is to be used by other researchers.
7. `empirical_evaluation`: Presenting novel empirical results discovered through experimentation, such as in running a model on a benchmark.
8. `analysis`: Presenting novel analyses of data that yield new insights.
9. `models_or_architectures`: Proposing completely novel model designs, or modifications to existing models/architectures.
10. `techniques_algorithms`: Introducing new learning algorithms, optimization techniques, or other algorithmic contributions.
11. `representational`: Novel ways to represent or encode data.
12. `research_methods_procedures`: Establishing novel experimental protocols, methods, or procedures for conducting aspects of research.
13. `metrics_instruments`: Developing novel evaluation metrics or protocols for assessment.
14. `position_statement`: Articulating a clear position on a debated topic, supported by evidence or argumentation.
15. `real_world_application`: Demonstrations that a research idea works in a practical, real-world setting.
16. `society_ethics_policy`: Addressing ethical, societal, and/or policy related issues.
17. `other`: Any other type of scientific contribution/claim not covered here.

When performing the classification, please use the above label exactly as it appears (e.g. `metrics_instruments`, not `metrics` or `instruments` or `metrics and instruments`).

# Contribution Granularity
When identifying contributions/claims, please identify them at a reasonable and relatively atomic level of granularity. A useful desideratum would be: Would a specific result be quoted as a central contribution/claim, or would it likely be broken down into a more atomic claim?
Example bad contributions (too coarse):
- 'evaluation of 3 models (A, B, C) on benchmark X': This should likely be broken down into multiple contributions/claims, one per model evaluated.
- 'set of 3 new metrics for evaluating Y': This should likely be broken down into multiple contributions/claims, one per metric.
But, also avoid being too fine-grained, especially if the results are intended to be taken together as a whole (and, are unlikely to be quoted individually later).
- `new optimization algorithm with 5 steps`: This is likely best kept as a single contribution/claim, unless the steps are independently useful contributions/claims on their own.
- `suite of 100 new benchmark tasks`: This is likely best kept as a single contribution/claim, unless the tasks are independently useful contributions/claims on their own.

**COMMON MISTAKE/LET'S BE CLEAR**: If the paper is testing a model on a task, that specific model/task combination should almost always be its own contribution, and not grouped with others -- unless (a) all the models are previously existing/unmodified, and (b) the results are only useful on aggregate, and not individually.
... (continued)
\end{lstlisting}
\end{figure*}

\lstset{firstnumber=last}   %
\begin{figure*}[t]
\begin{lstlisting}[language=json, caption={Example prompt for the contribution extraction subtask \textit{(part 2 of 2)}.}]
... (continued)
# Task Input
The input to this task is the full text of a research paper, delimited by triple backticks.  This paper text is provided above, between the === PAPER START === and === PAPER END === markers.

# Task Output
You are welcome to think as much as you feel appropriate before answering.  When you answer, please provide your answer in JSON format, with the following structure:
- A dictionary with a single key, `contributions`, which contains a list of contributions/claims
- Each contribution/claim is a dictionary with the following keys:
  - `name`: a short concise, information-dense name for the scientific contribution/claim made
  - `description`: a concise, information-dense, 3-5 sentence technical description of the scientific contribution/claim made. This is intended to (a) contain enough background context that it serves as a stand-alone description, and (b) contain enough detail that it clearly and precisely details the scientific contribution/claim.
  - `contribution_type`: a list of dictionaries, each with the following keys:
      - `type`: a string, showing one of the best-matching high-level types of contribution/claim made, from the list above.
      - `justification`: a concise, information-dense, 1 sentence technical justification for why this contribution/claim fits into this type.
  - `sections`: a list of strings, which exhaustively describe the sections in the paper where this contribution/claim is made or discussed.

For example:
```
{
  "contributions": [
    {
      "name": "Contribution 1",
      "description": "This paper presents Contribution 1, which is a novel approach to ...",
      "contribution_type": [
        {"type": "techniques_algorithms", "justification": "..."},
        {"type": "models_or_architectures", "justification": "..."}
      ],
      "sections": ["Introduction: Describes contribution 1", "Methods: Details of contribution 1"]
    },
    {
      "name": "Contribution 2",
      "description": "This paper also introduces Contribution 2, which improves upon ...",
      "contribution_type": [
        {"type": "empirical_evaluation", "justification": "..."},
      ],
      "sections": ["Results: Evaluation of contribution 2", "Discussion: Implications of contribution 2"]
    },
    ... # And so on, for all scientific contributions/claims made in the paper
  ]
}
```

# Description Details
Descriptions must be precise and technical, but must also contain enough background information to be stand-alone descriptions of the contributions/claims.
a) The context/background is required so that someone reading just the description will clearly understand what it is, what problem it applies to, and why it matters.  It will also (critically) be used to do search indexing, to find similar contributions.
b) The technical details are required so that the description is precise, detailed, and unambiguous about what the contribution/claim actually is.
Example:
- (bad) Integration of the three-part evaluation framework (task completion, procedural report-card, explanatory knowledge) to automatically grade each unit-test instance, providing fine-grained performance signals without human intervention.  -- Why bad: Lacks context/background about what the evaluation framework is, what problem it applies to, and so forth.
- (better) This contribution addresses the challenge of evaluating multi-step tasks in a scalable manner, by combining multiple evaluation dimensions into an automated grading system. The contribution is integrating the three-part evaluation framework for evaluating multi-step task performance (task completion, procedural report-card, explanatory knowledge) to automatically grade each instance of a series of multi-step unit tests, providing fine-grained performance signals without human intervention.  -- Why better: Provides broader context about the problem being solved, and allows more stand-alone understanding of what the contribution is about.  It's also more easily searchable, since it provides the context.

## Name Details
Similarly, the name must be concise, information dense, and descriptive.
Example:
- (bad) Curated hand-authored game list as a benchmark resource
- (better) Curated hand-authored *list of text games* as a benchmark resource *for evaluating agent models*  -- Why better: More specific and scoped to the contribution (what kind of games, what the purpose of the benchmark is)
# Instructions
You are welcome to think/reason as much as you feel appropriate before answering.  When you answer, please provide only the JSON output specified above. The JSON must be valid JSON output, and must be between triple backticks (```) so that it can be automatically parsed.
The format must be exactly as specified. That format again is:
```
{
  "contributions": [
    {
      "name": "...",
      "description": "...",
      "contribution_type": [
        {"type": "...", "justification": "..."},
        ... # And so on for all contribution types that clearly apply
      ],
      "sections": ["...", "...", ...]
    },
    ... # And so on for all contributions/claims made in the paper
  ]
}
```
You must be accurate, rigorous, truthful, and faithful to the ask.  You always act with the highest scientific integrity, and never make up information. Please pay particular attention to the contribution granularity guidelines above, to ensure the output is useful for technological roadmapping at a fine granularity. Your output should be ASCII formatted, unless Unicode is absolutely necessary (e.g. hyphens should be -, commas should be ', etc.). Do not hallucinate.
\end{lstlisting}
\end{figure*}

\lstset{firstnumber=1}   %
\begin{figure*}[t]
\begin{lstlisting}[language=json, caption={Example prompt for the prerequisite extraction subtask \textit{(part 1 of 4)}.}]
# Prerequisite Extraction Prompt 

You are an incredibly capable scientific research assistant. You always follow instructions carefully and accurately, and never make up information. You proceed with the highest scientific rigor and integrity.

Broad Task Context: Your broad task is in technological roadmapping: identifying the scientific contributions of research papers, and what prerequisite knowledge or capabilities were required to develop those scientific contributions.
The ultimate goal is to build a graph of scientific contributions, that clearly identifies how previous scientific contributions led to later scientific contributions, to better understand how making contributions today could lead to different kinds of capabilities in the future.

The task instructions come AFTER the paper. Treat them as the highest priority.

NOTE: The paper text was automatically extracted from the PDF, and may contain some extraction errors.
=== PAPER START ===
```text
<<VARIABLE: paper_text>>
```
=== PAPER END ===

=== TASK START ===

# Broad Task Context
Your broad task is in technological roadmapping: identifying the scientific contributions of research papers, and what prerequisite knowledge or capabilities were required to develop those scientific contributions.
The ultimate goal is to build a graph of scientific contributions, that clearly identifies how previous scientific contributions led to later scientific contributions, to better understand how making contributions today could lead to different kinds of capabilities in the future.

# Task
Your specific task is to read a research paper, and extract its scientific contributions/claims, and identify what prerequisite knowledge was required to make these contributions (nominally, with references to that prerequisite knowledge, if they are listed in the paper).
This will be a multi-step process:
Step 1: Read the paper carefully and identify its scientific contributions/claims
Step 2: For each contribution/claim, identify the prerequisite knowledge or capabilities required to make that contribution/claim
Step 3: For each prerequisite knowledge or capability, identify references to that prerequisite knowledge in the paper (if any)

Currently we are on *Step 2 and Step 3: Identify prerequisite knowledge or capabilities required to make a specific contribution/claim, and identify references to those prerequisites in the paper (if any)*

# Contribution/Claim Types
Here are some (non-exhaustive!) high-level types of scientific contributions/claims that are often made in research papers.  These may help guide your discovery/framing of scientific contributions/claims. Remember, this list is non-exhaustive, and other types of contributions/claims may exist.
When identifying contributions/claims, please classify each contribution/claim into one or more of these types (or `other` if none fit well):
1. `problem_formulation`: Reconceiving a task in a novel way, or identifying new angles/perspectives on long-standing problems. Can also be used to propose a new kind of task.
2. `theoretical_insight`: Proposing new theories, either from deriving theoretical results, or directly from empirical/analysis results.
3. `conceptual_framework`: Developing high-level conceptual frameworks that help guide future research.
4. `resource_benchmark`: Creating a new benchmark that evaluates model performance on a task.
5. `resource_dataset`: Creating new datasets that are primarily generic resources, without the intention of being used as a benchmark (i.e. without training/evaluation sets).
6. `tool_system_software`: Software, tools, systems, code, or other artifacts whose primary purpose is to be used by other researchers.
7. `empirical_evaluation`: Presenting novel empirical results discovered through experimentation, such as in running a model on a benchmark.
8. `analysis`: Presenting novel analyses of data that yield new insights.
9. `models_or_architectures`: Proposing completely novel model designs, or modifications to existing models/architectures.
10. `techniques_algorithms`: Introducing new learning algorithms, optimization techniques, or other algorithmic contributions.
11. `representational`: Novel ways to represent or encode data.
12. `research_methods_procedures`: Establishing novel experimental protocols, methods, or procedures for conducting aspects of research.
13. `metrics_instruments`: Developing novel evaluation metrics or protocols for assessment.
14. `position_statement`: Articulating a clear position on a debated topic, supported by evidence or argumentation.
15. `real_world_application`: Demonstrations that a research idea works in a practical, real-world setting.
16. `society_ethics_policy`: Addressing ethical, societal, and/or policy related issues.
17. `other`: Any other type of scientific contribution/claim not covered here.

When performing the classification, please use the above label exactly as it appears (e.g. `metrics_instruments`, not `metrics` or `instruments` or `metrics and instruments`).

# Contribution Granularity
When identifying contributions/claims, please identify them at a reasonable and relatively atomic level of granularity. A useful desideratum would be: Would a specific result be quoted as a central contribution/claim, or would it likely be broken down into a more atomic claim?
Example bad contributions (too coarse):
- 'evaluation of 3 models (A, B, C) on benchmark X': This should likely be broken down into multiple contributions/claims, one per model evaluated.
- 'set of 3 new metrics for evaluating Y': This should likely be broken down into multiple contributions/claims, one per metric.
But, also avoid being too fine-grained, especially if the results are intended to be taken together as a whole (and, are unlikely to be quoted individually later).
- `new optimization algorithm with 5 steps`: This is likely best kept as a single contribution/claim, unless the steps are independently useful contributions/claims on their own.
- `suite of 100 new benchmark tasks`: This is likely best kept as a single contribution/claim, unless the tasks are independently useful contributions/claims on their own.

**COMMON MISTAKE/LET'S BE CLEAR**: If the paper is testing a model on a task, that specific model/task combination should almost always be its own contribution, and not grouped with others -- unless (a) all the models are previously existing/unmodified, and (b) the results are only useful on aggregate, and not individually.

... (continued)
\end{lstlisting}
\end{figure*}

\lstset{firstnumber=last}   %
\begin{figure*}[t]
\begin{lstlisting}[language=json, caption={Example prompt for the prerequisite extraction subtask \textit{(part 2 of 4)}.}]
... (continued)

# Specific Contribution/Claim Being Analyzed
Previously, you identified the following scientific contribution/claim made in the paper (below).  Now, you must identify the prerequisite knowledge, capabilities, past results, or other scientific contributions that were required to make this contribution/claim.  Additionally, for each prerequisite identified, you must identify references to that prerequisite knowledge in the paper (if any -- some may not have references available).
```
<<VARIABLE: contribution (JSON)>>
```
NOTE: You may find that some contributions/claims need to be broken down to be more atomic, in order to properly identify their prerequisites.  If so, please do so, and output a set of contributions instead of just a more fully populated version of this single input contribution.

## For reference: Other contributions/claims identified in the paper
For reference, here are the other scientific contributions/claims you identified in the paper (below).  You may find that some of these contributions/claims serve as prerequisites for the contribution/claim being analyzed.
```
<<VARIABLE: other_contributions (JSON)>>
```

# Task Input
The input to this task is (1) A specific claim, shown above, and (2) the full text of a research paper, delimited by triple backticks.  The full text of the research paper is provided above, between the === PAPER START === and === PAPER END === markers.

# Task Output
You are welcome to think as much as you feel appropriate before answering.  When you answer, please provide your answer in JSON format, with the following structure:
- A dictionary with a single key, `contributions`, which contains a list of contributions/claims
- Each contribution/claim is a dictionary with the following keys:
(existing keys from input contribution)
  - `key`: a unique key for each contribution. This is already provided.
  - `name`: a short concise, information-dense name for the scientific contribution/claim made
  - `description`: a concise, information-dense, 3-5 sentence technical description of the scientific contribution/claim made. This is intended to (a) contain enough background context that it serves as a stand-alone description, and (b) contain enough detail that it clearly and precisely details the scientific contribution/claim.
  - `contribution_type`: a list of dictionaries, each with the following keys:
      - `type`: a string, showing one of the best-matching high-level types of contribution/claim made, from the list above.
      - `justification`: a concise, information-dense, 1 sentence technical justification for why this contribution/claim fits into this type.
  - `sections`: a list of strings, which exhaustively describe the sections in the paper where this contribution/claim is made or discussed.
(new keys for prerequisites)
  - `prerequisites`: a list of dictionaries, each with the following keys:
      - `name`: a short, concise, information-dense name for the prerequisite knowledge/capability required to make this contribution/claim
      - `description`: a concise, information-dense, 2-3 sentence technical description of the prerequisite knowledge/capability
      - `justification`: a concise, information-dense, 1-3 sentence technical justification for why this prerequisite knowledge/capability is required to make this contribution/claim
      - `core_or_peripheral`: a string, either `core` or `peripheral`, indicating whether this prerequisite knowledge/capability is core/central to making this contribution/claim, or peripheral/ancillary
      - `references_in_paper`: a list of dictionaries that provide references to this prerequisite knowledge/capability, if available from the paper. Each must be extracted exactly from the paper -- do not hallucinate references or try to generate them from your own knowledge. There are several possible formats for the references_in_paper entries:
          1) Internal references: for references that are other contributions/claims made in this same paper:
          - `type`: must be the string `internal`
          - `contribution_name`: (str) the name of the contribution/claim in this same paper that serves as a prerequisite (see the list of names of other contributions, above)
          - `contribution_key`: (str) the unique key of the contribution/claim in this same paper that serves as a prerequisite (see the list of other contributions, above)
          - `justification`: (str) a concise, information-dense, 1-3 sentence technical justification for why this other contribution/claim serves as a prerequisite
          2) Paper Reference: for references that are academic references (e.g. papers) -- this type is strongly preferred for external references:
          - `type`: must be the string `paper`
          - `paper_title`: (str) the title of the paper where this prerequisite came from
          - `first_author`: (dict) a dictionary describing the first author of that paper, with the keys 'last_name'(str), 'first_name'(str), and 'middle_names'(str)
          - `year`: (int) the year the paper was published
          - `venue`: (str) the venue where the paper was published
          - `corpus_id`: (int) the S2 corpus ID of the paper the prerequisite came from, if available. Nominally, available in the `paper_corpus_id` key in the reference list. Use `None` if not available.
          3) Other references: for references that are other types (e.g. software, etc.) that do NOT have an accompanying academic reference, but DO have a URL:
          - `type`: must be the string `other`
          - `name`: (str) a short name describing the reference
          - `url`: (str) the URL where this resource can be found

For example:
```
{
  "contributions": [
    {
      "key": "...",
      "name": "Contribution 1",
      "description": "This paper presents Contribution 1, which is a novel approach to ...",
      "contribution_type": [
        {"type": "techniques_algorithms", "justification": "..."},
        {"type": "models_or_architectures", "justification": "..."}
      ],
      "sections": ["Introduction: Describes contribution 1", "Methods: Details of contribution 1"],

... (continued)
\end{lstlisting}
\end{figure*}

\lstset{firstnumber=last}   %
\begin{figure*}[t]
\begin{lstlisting}[language=json, caption={Example prompt for the prerequisite extraction subtask \textit{(part 3 of 4)}.}]
... (continued)

      "prerequisites": [
        {
          "name": "Prerequisite 1",
          "description": "Prerequisite 1 is a method for ...",
          "justification": "This prerequisite is required because ...",
          "core_or_peripheral": "core",
          "references_in_paper": [
            {
              "type": "paper",
              "paper_title": "Title of the prerequisite paper",
              "first_author": {"last_name": "Smith", "first_name": "John", "middle_names": "A."},
              "year": 2020,
              "venue": "Journal of Important Results",
              "corpus_id": 123456789
            },
            {
              "type": "internal",
              "contribution_name": "Contribution 2",
              "contribution_key": "2",
              "justification": "..."
            },
            {
              "type": "other",
              "name": "Software Library X",
              "url": "https://example.com/library_x"
            },
            ... # And so on for all references to this prerequisite in the paper. The above are just examples; use any combination of `paper`, `internal`, and/or `other`, including multiple instances of one (e.g. multiple papers, and/or multiple internals, etc).
          ]
        },
        ... # And so on, for all prerequisites for this contribution
    },
    ... # If you are breaking down the single input contribution into more atomic contributions, please add more contributions here
  ]
}
```

## Contribution keys
The contribution key (`key`) should ALWAYS be the same as the input contributions key, and formatted as a string, even if it is numeric.  If you split this contribution into multiple contributions, add a dash to the new contributions (e.g. input key `2` would be split into `2-1`, 2-2`, `2-3`, and so on).

# Description Details
Descriptions must be precise and technical, but must also contain enough background information to be stand-alone descriptions of the contributions/claims.
a) The context/background is required so that someone reading just the description will clearly understand what it is, what problem it applies to, and why it matters.  It will also (critically) be used to do search indexing, to find similar contributions.
b) The technical details are required so that the description is precise, detailed, and unambiguous about what the contribution/claim actually is.
Example:
- (bad) Integration of the three-part evaluation framework (task completion, procedural report-card, explanatory knowledge) to automatically grade each unit-test instance, providing fine-grained performance signals without human intervention.  -- Why bad: Lacks context/background about what the evaluation framework is, what problem it applies to, and so forth.
- (better) This contribution addresses the challenge of evaluating multi-step tasks in a scalable manner, by combining multiple evaluation dimensions into an automated grading system. The contribution is integrating the three-part evaluation framework for evaluating multi-step task performance (task completion, procedural report-card, explanatory knowledge) to automatically grade each instance of a series of multi-step unit tests, providing fine-grained performance signals without human intervention.  -- Why better: Provides broader context about the problem being solved, and allows more stand-alone understanding of what the contribution is about.  It's also more easily searchable, since it provides the context.

## Name Details
Similarly, the name must be concise, information dense, and descriptive.
Example:
- (bad) Curated hand-authored game list as a benchmark resource
- (better) Curated hand-authored *list of text games* as a benchmark resource *for evaluating agent models*  -- Why better: More specific and scoped to the contribution (what kind of games, what the purpose of the benchmark is)

# Instructions
You are welcome to think/reason as much as you feel appropriate before answering.  When you answer, please provide only the JSON output specified above. The JSON must be valid JSON output, and must be between triple backticks (```) so that it can be automatically parsed.
The format must be exactly as specified. That format again is:
```
{
  "contributions": [
    {
      "key": "...",
      "name": "...",
      "description": "...",
      "contribution_type": [
        {"type": "...", "justification": "..."},
        ... # And so on for all contribution types that clearly apply
      ],
      "sections": ["...", "...", ...],

... (continued)
\end{lstlisting}
\end{figure*}

\lstset{firstnumber=last}   %
\begin{figure*}[t]
\begin{lstlisting}[language=json, caption={Example prompt for the prerequisite extraction subtask \textit{(part 4 of 4)}.}]
... (continued)

      "prerequisites": [
        {
          "name": "...",
          "description": "...",
          "justification": "...",
          "core_or_peripheral": "...",
          "references_in_paper": [
            {
              "type": "paper",
              "paper_title": "...",
              "first_author": {"last_name": "...", "first_name": "...", "middle_names": "..."},
              "year": ...,
              "venue": "...",
              "corpus_id": ...
            },
            ... # And so on for all references to this prerequisite in the paper. Could also use the `internal` or `other` type reference.
          ]
        },
        ... # And so on, for all prerequisites for this contribution
    },
    ... # If you are breaking down the single input contribution into more atomic contributions, please add more contributions here
  ]
}
```

You must be accurate, rigorous, truthful, and faithful to the ask.  You always act with the highest scientific integrity, and never make up information.
Please pay particular attention to the contribution granularity guidelines above, to ensure the output is useful for technological roadmapping at a fine granularity.
Your output should be ASCII formatted, unless Unicode is absolutely necessary (e.g. hyphens should be -, commas should be ', etc.)
Do not hallucinate.
=== TASK END ===

... (continued)
\end{lstlisting}
\end{figure*}

\lstset{firstnumber=1}   %
\begin{figure*}[t]
\begin{lstlisting}[language=json, caption={Example prompt for the cross-paper prerequisite-to-contribution alignment step.}]
# Cross-paper Prerequisite-to-Contribution Alignment Prompt

You are ScientistGPT, an expert AI scientist. You can answer any scientific problem correctly, faithfully, and accurately, using the highest scientific integrity.

# Task
You are building a graph of scientific contributions (extracted from papers) and what scientific contributions were required as prerequisites for each contribution. The goal is to make a graph that links a specific contribution in Paper 1 (say, from 2025) to one or more contributions from other papers that enabled that contribution.
Previously, you extracted a list of scientific contributions (and their prerequisites) from a set of papers.
Now, you will be shown:
1) Source Paper: A specific scientific contribution (with description) from a paper, and a specific prerequisite.
2) Cited Paper: A list of scientific contributions from one of the papers it cited as a prerequisite.
Your task is to determine which of the contributions from the cited paper match the prerequisite from the source paper.

# Source Paper Information: Specific Contribution and Prerequisites
Here is the specific contribution and prerequisite from the source paper:
```
<<VARIABLE: source_contribution_with_prerequisite (JSON)>>
```

# Cited Paper Information: List of Contributions
Here is the list of contributions extracted from the cited paper:
```
<<VARIABLE: cited_paper_record (JSON)>>
```

# Instructions
Based on the information above, identify which contributions from the cited paper match the prerequisite from the source paper.
It is possible that one, multiple, or none of the contributions from the cited paper match the prerequisite.

## Desiderata for determining if there is a strong match
- The contribution from the cited paper should directly address the scientific concept, method, or technology described in the prerequisite from the source paper.
- The contribution from the cited paper should provide a clear foundation or basis that enables the advancement described in the prerequisite.
If the contribution from the cited paper meets these criteria, then it is considered a `strong` match.
If the contribution from the cited paper only tangentially relates to the prerequisite, then it is considered a `weak` match.
If none of the contributions from the cited paper adequately address the prerequisite, then there is no match.

# Response Format
You are strongly encouraged to think as much as appropriate before answering.
You must respond in JSON format, and your JSON must be valid JSON, and between codeblocks (```).
Your JSON response must be a dictionary with the following keys:
- `matches`: A list of dictionaries, each with the following keys:
  - `contribution_key`: The `key` of the contribution from the cited paper that matches the prerequisite. This must match exactly, and be in the same type (e.g. a string to a string).
  - 'explanation': A brief (1-2 sentence) explanation of why this contribution matches the prerequisite.
  - `match_type`: Either `strong` or `weak`, indicating the strength of the match    
- `overall_explanation`: A brief (1-2 sentence) explanation summarizing the overall matching process and results.
If there are no matches, then the `matches` list should be empty (`[]`), and `overall_explanation` should explain why no matches were found.

## Example Response
The following is a toy example of a valid JSON response:
```
{
    "matches": [
        {
            "contribution_key": "3",
            "explanation": "...",
            "match_type": "strong"
        },
        # Add more matches as needed
    ],
    "overall_explanation": "..." 
}
```

Please provide your response below.
You must be accurate, rigorous, truthful, and faithful to the ask.  You always act with the highest scientific integrity, and never make up information.
Please pay particular attention to the guidelines above, to ensure the output is useful for technological roadmapping at a fine granularity.
Your output should be ASCII formatted, unless Unicode is absolutely necessary (e.g. hyphens should be -, commas should be ', etc.)
Do not hallucinate.
\end{lstlisting}
\end{figure*}

\end{document}